\documentclass[letterpaper, 10 pt, conference]{ieeeconf}

\IEEEoverridecommandlockouts                              %

\usepackage{xcolor}
\usepackage{amsmath}
\usepackage{amssymb}
\usepackage{multirow}
\usepackage{booktabs}
\usepackage{graphicx}
\usepackage{mdframed}
\usepackage{caption}
\usepackage{verbatim}
\usepackage{xspace}
\usepackage{paralist}
\usepackage{cuted}
\usepackage{hyperref}
\hypersetup{
    colorlinks=true,
    linkcolor=blue,
    filecolor=magenta,      
    urlcolor=cyan,
}
\usepackage{url} 
\usepackage{algorithm}
\usepackage{algpseudocode}
\usepackage{tcolorbox}
\tcbuselibrary{breakable}

\usepackage{listings}
\usepackage{xcolor}
\usepackage{courier} %
\usepackage[noadjust]{cite}

\newcommand{\figref}[1]{Figure~\ref{#1}}

\newcommand{\algname}{Iterative Keypoint Reward\xspace}
\newcommand{\algabrvname}{IKER\xspace}

\title{\LARGE \bf
A Real-to-Sim-to-Real Approach to Robotic Manipulation\\with VLM-Generated Iterative Keypoint Rewards
}

\author{Shivansh Patel$^{1*}$, Xinchen Yin$^{1*}$, Wenlong Huang$^{2}$, Shubham Garg$^{3}$, Hooshang Nayyeri$^{3}$, \\ Li Fei-Fei$^{2}$, Svetlana Lazebnik$^{1}$, Yunzhu Li$^{4}$ %
\thanks{*indicates equal contribution. $^{1}$University of Illinois at Urbana-Champaign, $^{2}$Stanford University, $^{3}$Amazon, $^{4}$Columbia University}%
}

\begin{document}

\newcommand{\todo}[1]{{\color{red}{TODO: }#1}}
\newcommand{\SP}[1]{{\color{blue}{SP: }#1}}
\newcommand{\XY}[1]{{\color{brown}{XY: }#1}}
\newcommand{\WH}[1]{{\color{Sepia}{WH: }#1}}
\newcommand{\YL}[1]{{\color{purple}{YL: }#1}}
\newcommand{\SL}[1]{{\color{orange}{SL: }#1}}

\newcommand{\updated}[1]{{\color{cyan}#1}}

\newcommand{\xhdr}[1]{\vspace{2pt}\noindent\textbf{#1}}
\newcommand{\etal}{\textit{et al.}}

\maketitle
\thispagestyle{empty}
\pagestyle{empty}
\begin{abstract}
Task specification for robotic manipulation in open-world environments is challenging, requiring flexible and adaptive objectives that align with human intentions and can evolve through iterative feedback. We introduce \algname (\algabrvname), a visually grounded, Python-based reward function that serves as a dynamic task specification. Our framework leverages VLMs to generate and refine these reward functions for multi-step manipulation tasks. Given RGB-D observations and free-form language instructions, we sample keypoints in the scene and generate a reward function conditioned on these keypoints. IKER operates on the spatial relationships between keypoints, leveraging commonsense priors about the desired behaviors, and enabling precise SE(3) control. We reconstruct real-world scenes in simulation and use the generated rewards to train reinforcement learning (RL) policies, which are then deployed into the real world—forming a real-to-sim-to-real loop. Our approach demonstrates notable capabilities across diverse scenarios, including both prehensile and non-prehensile tasks, showcasing multi-step task execution, spontaneous error recovery, and on-the-fly strategy adjustments. The results highlight \algabrvname's effectiveness in enabling robots to perform multi-step tasks in dynamic environments through iterative reward shaping. Project Page: \url{https://iker-robot.github.io/}
\end{abstract}

\section{Introduction}

\begin{figure}[t]
    \centering
    \setlength{\abovecaptionskip}{-10pt}
    \includegraphics[width=\linewidth]{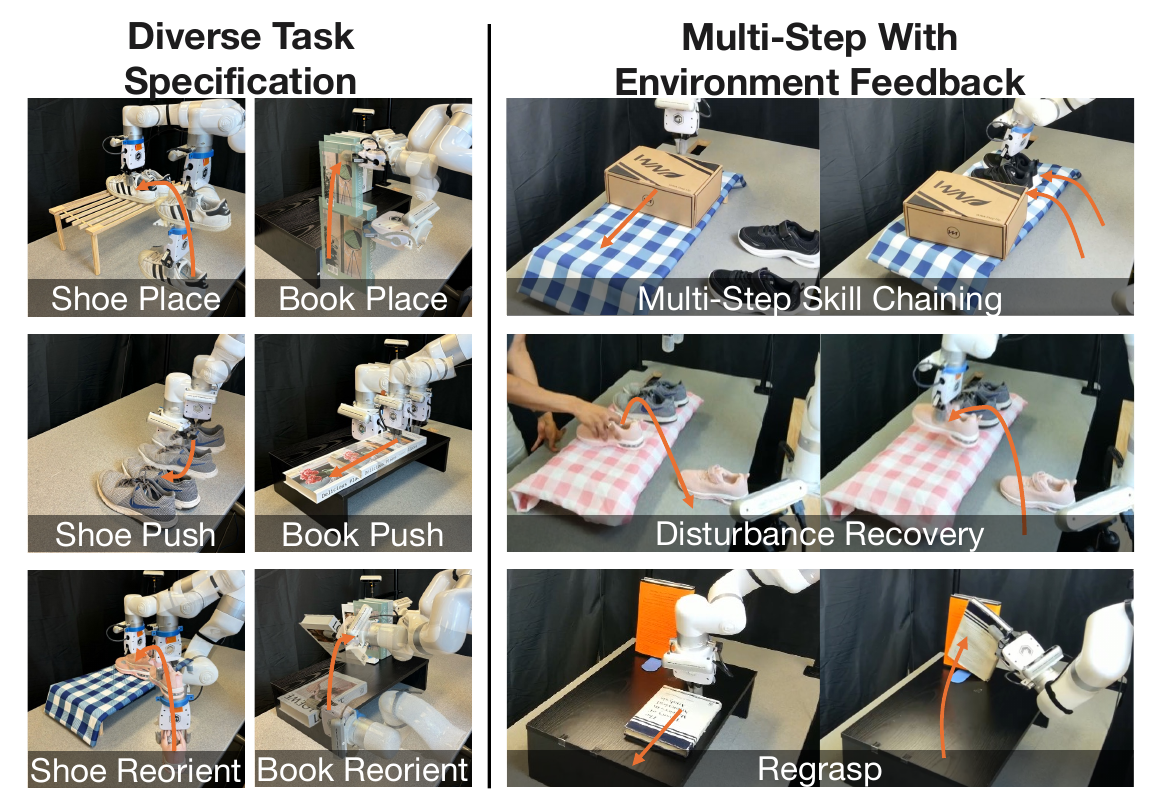}
    \captionsetup{type=figure}
    \caption{\small{\textbf{Capabilities of Our Framework.} IKER is designed to handle a wide range of real-world tasks. It can be seamlessly chained to execute multi-step tasks. It exhibits robustness to disturbances and demonstrates the ability to solve problems flexibly.}}
    \label{fig:teaser}
\end{figure}

Suppose that a robot is tasked with placing a pair of shoes on a rack, but a shoe box is occupying the rack, leaving insufficient space for both shoes (\figref{fig:teaser}, top right). The robot must first push the box aside to create space and then proceed to place the shoes. This example highlights the importance of task specification for robots in unstructured, real-world environments, where tasks can often involve multiple implicit steps. In such cases, rigid predefined instructions fail to capture the complexities of interaction required to accomplish the goal. To be effective, task specifications must incorporate commonsense priors—expectations about how the robot should behave. For instance, rather than attempting to squeeze the shoes in awkwardly, the robot should realize that it must first clear space.

Recent vision-language models (VLMs) show promise for freeform robotic task specification due to their rapidly advancing ability to encode rich world knowledge by pretraining on vast and diverse datasets~\cite{openai2023gpt,zeng2022socratic, radford2021learningtransferablevisualmodels, jia2021scalingvisualvisionlanguagerepresentation, li2022blip, li2023blip2bootstrappinglanguageimagepretraining, alayrac2022flamingovisuallanguagemodel, yu2022cocacontrastivecaptionersimagetext}. VLMs excel in interpreting natural language descriptions and complex instructions. Their broad knowledge bridges human expectations and robot behavior, capturing human-like priors and problem-solving strategies. However, previous works that leverage VLMs in robotics face two major limitations: (1) they lack the capability to specify precise target locations in 3D, and (2) they are often unable to adapt to the environment changes as the task progresses.

In this work, we introduce \textbf{\algname~(\algabrvname)}, a visually grounded reward function for robotic manipulation that addresses these limitations. %
Inspired by recent work~\cite{huang2024rekep}, we draw the observation that both object positions and orientations can be encoded using keypoints. Hence, \algabrvname allows for fine-grained manipulation in 3D, facilitating complex tasks that require accurate location and orientation control. Additionally, \algabrvname incorporates an iterative refinement mechanism, where the VLM updates the task specification based on feedback from the robot's interactions with the environment. This mechanism enables %
dynamically-adjusting strategies and intermediate steps, such as repositioning objects for a better grasp.

While VLMs excel in processing real-world visual data, training policies directly in the real world is often infeasible due to safety, scalability, and efficiency constraints. To address this, we first generate \algabrvname using real-world observations, then transfer the scene and the reward to simulation for training, and finally, deploy the optimized policy back into the real world. Thus, our system operates in a real-to-sim-to-real loop. %

We demonstrate the efficacy of our real-to-sim-to-real framework with \algabrvname across diverse scenarios involving everyday objects like shoes and books. These include prehensile tasks, such as placing shoes on racks, and non-prehensile tasks, like sliding books to target locations. We conduct both quantitative and qualitative evaluations to assess the system's ability to perform complex, long-horizon tasks autonomously. The results showcase human-like capabilities, including multi-step action sequencing, spontaneous error recovery, and the ability to update strategies in response to changes in the environment.

\begin{figure*}[t]
    \setlength{\abovecaptionskip}{-12pt}
    \centering
    \includegraphics[width=\linewidth]{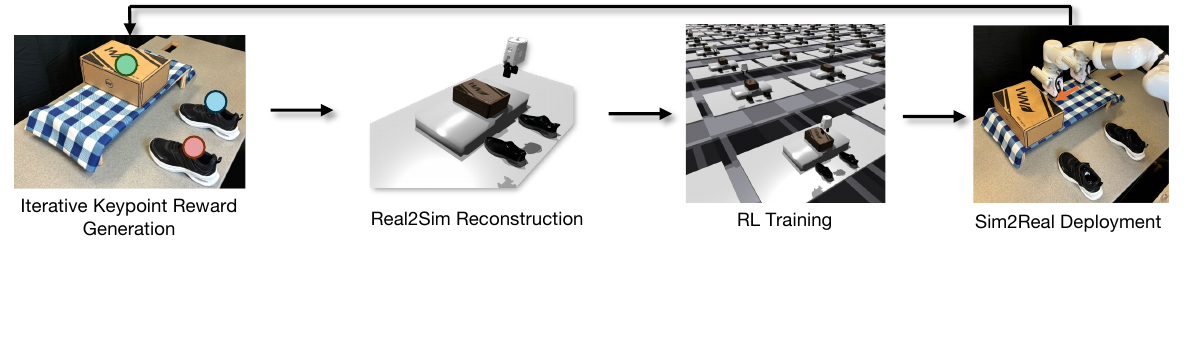}
    \vspace{-4.8em} %
    \captionsetup{type=figure}
    \caption{\small{\textbf{Framework Overview.} \algname (\algabrvname) is a visually grounded reward generated by Vision-Language Models (VLMs) as task specification. The framework reconstructs the real-world scene in simulation, and the generated reward is used to train RL policies, which are subsequently deployed in the real-world.}}
    \label{fig:overview}
    \vspace{-1.5em}
\end{figure*}

\section{Related Work}
\xhdr{VLMs in Robotics.} VLMs have become a prominent tool in robotics~\cite{ahn2022can,codeaspolicies2022,huang2023voxposer,brohan2022rt,brohan2023rt,liu2024moka,huang2024copa,team2024octo,huang2023instruct2act,xu2023creative,zhou2023generalizable,nasiriany2024pivot,di2024keypoint,zeng2023large,zha2024distilling,arenas2024prompt,mahadevan2024generative,liang2024learning,huang2024grounded,ren2023robots,jiang2022vima,yang2024guiding,duan2024aha,duan2024manipulate,yuan2024robopoint,singh2023progprompt,tang2024kalie,liang2024eurekaverse,zawalski2024robotic}. Existing works utilizing VLMs in robotics primarily focus on two areas: task specification~\cite{ahn2022can,codeaspolicies2022,huang2023voxposer,liu2024moka,huang2024copa,nasiriany2024pivot} and low-level control~\cite{brohan2022rt,brohan2023rt,o2023open,codeaspolicies2022}. Our work aligns with the former, with an emphasis on flexibility and adaptability in complex, real-world environments.

For task specification, many works employ VLMs to break down complex tasks into manageable subtasks, demonstrating their utility in bridging high-level instructions and robotic actions. Huang \etal~\cite{huang2022language} demonstrate the use of LLMs as zero-shot planners, enabling task decomposition into actionable steps. Similarly, Ahn \etal~\cite{ahn2022can} leverage VLMs to parse long-horizon tasks and sequence them into executable steps for robots. Belkhale \etal~\cite{rth2024arxiv} introduce ``language motions'' that serve as intermediaries between high-level instructions and specific robotic actions, allowing policies to capture reusable, low-level behaviors.  Unlike these works, our approach focuses on flexible interpretation of tasks in the context of a dynamically changing environment.

Beyond task decomposition, VLMs have been used to generate affordances and value maps that guide robotic actions. Huang \etal~\cite{huang2023voxposer} employs VLMs to generate 3D affordance maps, providing robots with spatial knowledge of which parts of the environment are suitable for interaction. Liu \etal~\cite{liu2024moka} use VLMs to predict point-based affordances, enabling zero-shot manipulation tasks. Zhao \etal~\cite{zhao2024vlmpc} incorporate VLMs into model predictive control, where the models predict the outcomes of candidate actions to guide optimal decision-making. These works demonstrate the potential of VLMs to bridge high-level task understanding with spatial and functional knowledge needed for robotic control. Similar to our work, Huang \etal~\cite{huang2024rekep} use keypoints and define relations and constraints between them to execute manipulation tasks, but their approach follows an open-loop strategy. In contrast, we employ a closed-loop approach, enabling dynamic plan adjustments. Additionally, our approach also supports non-prehensile manipulations, such as pushing.

Some works have also explored VLMs for reward function generation~\cite{yu2023language, ma2023eureka, ma2024dreurekalanguagemodelguided,xie2023text2reward}. However, most of these approaches have limited real-world applicability. Some lack demonstrations on real robots~\cite{ma2023eureka}, are restricted to a single real-world scenario~\cite{xie2023text2reward}, or focus on highly constrained tasks like a robot dog walking on a ball~\cite{ma2024dreurekalanguagemodelguided}. In contrast, our work demonstrates the versatility and robustness of VLM-generated rewards on multiple real-world manipulation tasks.

\xhdr{Real-to-Sim and Sim-to-Real.}
Real-to-sim has gained significant attention for its ability to facilitate agent training. Once a scene is transferred to simulation, it can be used for a wide range of tasks, including RL. Several approaches focus on reconstructing rigid bodies for use in simulation~\cite{kappler2018real,wen2022you,liu2024one,xu2025sparp,shi2023zero123++,liu2023zero,gao2022get3d}. For instance, Kappler et al.~\cite{kappler2018real} introduce a method for reconstructing rigid objects to facilitate grasping. 
Some works rather focus on reconstructing articulated objects~\cite{mu2021sdf,jiang2022ditto,nie2022structure,chen2024urdformer,mandi2024real2code,liu2023paris,liu2024cage,liu2024singapo}. Huang et al.~\cite{huang2012occlusion} present methods for reconstructing the occluded shapes of articulated objects. Jiang \etal~\cite{jiang2022ditto} introduce a framework, DITTO, to generate digital twins of articulated objects from real-world interactions. In our work, we utilize the fast state-of-the-art BundleSDF method~\cite{wen2023bundlesdf} to generate object meshes that are transferred to the simulation. 

Sim-to-real transfer has shown great performance in a variety of skills, including tabletop manipulation~\cite{shridhar2021cliport, jiang2024transicsimtorealpolicytransfer}, mobile manipulation~\cite{gu2022multiskill, yenamandra2023homerobot}, dynamic manipulation~\cite{huang2023dynamic}, dexterous manipulation~\cite{chen2023sequential, qin2022dexpointgeneralizablepointcloud,qi2023hand,yin2023rotating}, and locomotion~\cite{DBLP:conf/rss/KumarFPM21,he2024agilesafelearningcollisionfree}. 
However, directly deploying learned policies to physical robots cannot guarantee successful performance due to the sim-to-real gap. 
To bridge this gap, researchers have developed many techniques, such as system identification~\cite{tan2018simtoreal, chang2020sim2real2sim, lim2021planar}, domain adaptation~\cite{bousmalis2018using, arndt2019metareinforcementlearningsimtoreal, rao2020rlcycleganreinforcementlearningaware,james2019sim,du2022bayesian}, and domain randomization~\cite{DBLP:conf/rss/KumarFPM21, openai2019solving, tobin2017domain,antonova2021bayessimig,lee2020learning,peng2018sim,tobin2017domain,chebotar2019closing}. 
In our work, we use domain randomization as it does not require any interaction data from the real world during training. It relies entirely on simulation and makes policies robust by exposing them to a wide variety of randomized conditions. Recently, Torne \etal~\cite{torne2024reconciling} proposed RialTo, a complete real-to-sim-to-real loop system that focuses on leveraging simulation to robustify imitation learning policies trained using real-world collected demonstrations. In contrast, we focus on executing long-horizon tasks by training only in simulation, bypassing the need for demonstrations.

\section{Method}

\begin{algorithm}[t]
\scriptsize
\begin{algorithmic}[1]
    \raggedright
    \State \textbf{Given: } Language instruction $I$
    \State \textbf{Initialize:} \texttt{done} $\gets$ \texttt{false}, \texttt{execution\_history} $\gets$ [\ ], $i \gets 1$
    \State Generate 3D models of objects
    \While{\texttt{true}}
        \State $(\{k_{j}^{(i)}\}_{j=1}^{K_i}, O_i) \gets$ \texttt{GetKeypoints}(3D models)
        \State \texttt{code} $\gets$ \texttt{QueryVLM}($O_i$, \texttt{execution\_history})
        \State $(\texttt{done}, \{k_{j}^{\text{target}(i)}\}_{j=1}^{K_i}) \gets$ \texttt{Execute}(\texttt{code})
        \If{\texttt{done} is \texttt{true}}
            \State \texttt{break}
        \EndIf
        \State $s_i \gets$ \texttt{TransferSceneToSimulation}(3D models)
        \State $\pi_i \gets$ \texttt{LearnPolicy}($s_i$, $\{k_{j}^{\text{target}(i)}\}_{j=1}^{K_i}$)
        \State \texttt{ExecutePolicyInRealWorld}($\pi_i$)
        \State Append ($O_i$, \texttt{code}) to \texttt{execution\_history}
        \State $i \gets i + 1$
    \EndWhile
\end{algorithmic}
\caption{IKER Execution Framework}
\label{alg:vlm_execution}
\end{algorithm}

In this section we formally define \algname{} (\algabrvname{}) and discuss how it is automatically synthesized and refined by VLMs by continuously taking in environmental feedback. Then, we discuss our overall framework, which uses \algabrvname in a real-to-sim-to-real loop. Our method overview is illustrated in \figref{fig:overview}, with detailed steps provided in Algorithm ~\ref{alg:vlm_execution}.

\subsection{\algname~(\algabrvname)}

\begin{figure}[t]
    \setlength{\abovecaptionskip}{0pt}
    \includegraphics[width=\linewidth]{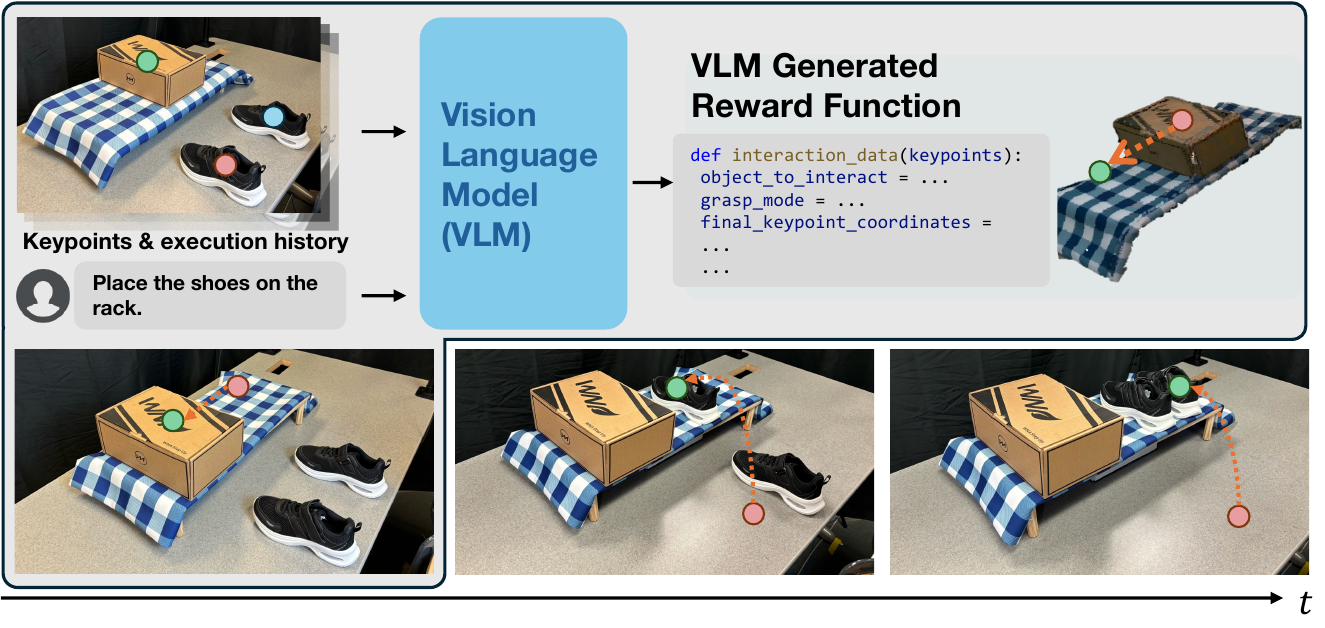}
    \caption{\small{\textbf{\algname Generation.} This corresponds to the first step in \figref{fig:overview}. We first obtain keypoints in the scene. These keypoints, combined with a human command and execution history, are processed by a VLM to generate code that maps keypoints to the reward function.  A more detailed illustration of the keypoints and generated code is provided in \figref{fig:unrolled}.}}
    \label{fig:method}
    \vspace{-1.5em}
\end{figure}

Given an RGB-D observation of the environment and an open-vocabulary language instruction $I$ for a multi-step task, our goal is to obtain a sequence of policies, $\pi_{i=1}^{N}$, that complete the task. Crucially, the number of policies $N$ is not predefined, allowing for flexibility in how the robot approaches the task. For example, in the scenario of Fig.~\ref{fig:method}, the first policy, $\pi_1$, moves the shoe box to create space, while subsequent policies handle the placement of each shoe.

For each step $i$, we denote the RGB observation as $O_i$. We assume a set of $K_i$ keypoints $\{k_j^{(i)}\}_{j=1}^{K_i}$ is given (discussed later in Sec.~\ref{sec:real2sim}), each specifying a 3D position in the task space. Using these keypoints, our objective is to automatically generate a reward function, termed \algabrvname, that maps the keypoint positions to a scalar reward $f^{(i)}: \mathbb{R}^{K_i \times 3} \rightarrow \mathbb{R}$.

To generate the reward function $f^{(i)}$, we use a VLM (GPT-4o~\cite{openai2023gpt} in our case), which is provided with the context comprising (1) the human instruction $I$ describing the task, (2) the current RGB observation $O_i$ with keypoints overlaid with numerical markers, and (3) the sequence of previous observations and reward functions up to step $i - 1$, i.e. $\{O_1, f^{(1)}, \dots, O_{i-1}, f^{(i-1)}\}$. 
 
Additionally, the VLM is guided by a prompt that instructs it to generate a Python function for the reward $f^{(i)}$. The prompt directs the VLM to break down the task into executable steps, predict which object to interact with, specify the movement of objects by indicating where their keypoints should be placed relative to other keypoints, and perform arithmetic calculations on these keypoints to predict their final locations. We do not explicitly specify which keypoint belongs to which object, allowing the VLM to infer this information. The prompt also instructs the VLM to present all outputs in a prescribed code format and set the flag \texttt{done = True} if the task is completed. By predicting code, the VLM can perform arbitrary and precise calculations using the current keypoint locations, which would not be possible if limited to raw text. Please refer to Appendix~\ref{sec:prompt} for the complete prompt, and \figref{fig:unrolled} for a step-by-step walkthrough of RGB observations and generated reward functions for the example of \figref{fig:method}.

Upon receiving the final keypoint locations by executing the generated code, we compute a scalar reward to evaluate the policy's performance. The reward function, $f^{(i)}$, facilitates learning by combining the following terms:

\begin{compactitem} 
    \item Gripper-to-object Distance Reward ($r_{\text{dist}}$): Encourages the robot to approach the object of interest by penalizing large distances between them. 
    \item Direction Reward ($r_{\text{dir}}$): Guides the robot to move the keypoints in the direction of the target locations. 
    \item Alignment Reward ($r_{\text{align}}$): Drives the robot to position the keypoints close to their target locations.
    \item Success Bonus ($r_{\text{bonus}}$): Provides an additional reward when the average distance between the keypoints and their target positions remains within a specified threshold for a certain number of timesteps, indicating successful task completion. 
    \item Penalty Term ($r_{\text{penalty}}$): Applies penalties for undesirable actions such as excessive movements, dropping the object, or applying excessive force.
\end{compactitem}

\vspace{-5mm}
\[
\scalebox{0.85}{$
f^{(i)} = \alpha_{\text{dist}} r_{\text{dist}} + \alpha_{\text{dir}} r_{\text{dir}} + \alpha_{\text{align}} r_{\text{align}} + \alpha_{\text{bonus}} r_{\text{bonus}} + \alpha_{\text{penalty}} r_{\text{penalty}}
$}
\]
\vspace{-5mm}

\subsection{Transferring real-world scene to simulation}\label{sec:real2sim}
To transfer the real-world scene within the workspace boundary to simulation, we first generate 3D meshes of manipulable objects, such as the shoe box and shoes shown in \figref{fig:method}, by capturing video footage of each object as it is moved to ensure the camera captures all sides. These videos allow for accurate 3D mesh reconstruction using BundleSDF~\cite{wen2023bundlesdf}, and multiple objects can be processed in parallel to speed up the scanning phase. Once a mesh is created for an object, it can be reused in different settings, eliminating the need to recreate it for each new scenario. With the meshes prepared, we use FoundationPose~\cite{wen2023foundationpose} to estimate the objects' poses, enabling precise placement of the corresponding meshes in the simulated environment. For static elements, like the workspace table and shoe rack in \figref{fig:method}, we capture a point cloud to create their meshes for use in the simulation.

The generated meshes are further used to identify candidate keypoints. For manipulable objects like shoes or books, keypoints are placed at the object's extremities along its axes, defined with respect to the object's center, independent of the human instruction. For static objects like shoe racks, which are part of the environment, keypoints are uniformly distributed across their surfaces. Numerical labels assigned to keypoints are grouped by objects. For example, as shown in \figref{fig:unrolled}, keypoints 1–4 correspond to the box, 5–8 to the left shoe, 9–12 to the right shoe, and the remaining keypoints to the rack. Keypoints that are too close together in the image projection are removed. Specifically, background keypoints (e.g., like rack) near object keypoints are removed first. Among overlapping object keypoints, only the one with the lower numerical label is retained. Note, however, that the VLM is not explicitly told this information but has to infer the association between the keypoints and the objects based on the input image.

\subsection{Policy Training in Simulation}
\label{sec:training}
We control the robot in the end-effector space, which has six degrees of freedom: three prismatic joints for movement along the x, y, and z axes, and three revolute joints for rotation.
The gripper fingers remain closed by default, opening only when grasping objects. Refer to Appendix~\ref{sec:grasping} for a detailed discussion on grasping. 

\textbf{State Space}: 
The state space for our policy captures the essential information to execute the task. The input is a vector \( s_t \) consisting of the gripper's end-effector pose \( (\mathbf{p}_e, \mathbf{q}_e) \in \mathbb{R}^7 \), the pose of object currently being manipulated \( (\mathbf{p}_o, \mathbf{q}_o) \in \mathbb{R}^7 \), a set of object keypoints \( \mathcal{K}_o = \{k_j^{(i)}\}_{j=1}^{K_i} \in \mathbb{R}^{K_i \times 3} \), and their corresponding target positions \( \mathcal{K}_t = \{k_{t_j}^{(i)}\}_{j=1}^{K_i} \in \mathbb{R}^{K_i \times 3} \). $\mathcal{K}_o$ is calculated by applying rigid body transformations to keypoints defined in the object's local coordinate frame, mapping them to their corresponding positions in the world frame. $\mathcal{K}_t$ is derived from the reward function $f^{(i)}$ generated by the VLM. Rotations $\mathbf{q}_e$ and $\mathbf{q}_o$ are represented as quaternions. 
This state space \( s_t = (\mathbf{p}_e, \mathbf{q}_e, \mathbf{p}_o, \mathbf{q}_o, \mathcal{K}_o, \mathcal{K}_t) \) captures essential information on objects of interest as well as the goal of the policy. Instead of incorporating raw RGBD data directly into the state space, object poses and keypoints are extracted from RGBD inputs using a vision-based pose estimation method, as detailed in Section~\ref{sec:deployment}. This preprocessing step removes the necessity of including raw RGBD data in the policy.

\textbf{Action Space}: The action space is defined relative to the gripper's current position and orientation. The policy outputs actions \( a_t = (\Delta \mathbf{p}_e, \Delta \mathbf{r}_e) \), where \( \Delta \mathbf{p}_e \in \mathbb{R}^3 \) and \( \Delta \mathbf{r}_e \in \mathbb{R}^3 \) specifies the changes in translation and rotation respectively.

\textbf{Training Algorithm \& Architecture}: We train our policies using IsaacGym~\cite{makoviychuk2021isaac} simulator with the PPO~\cite{schulman2017proximal} algorithm. We use an actor-critic architecture~\cite{konda1999actor} with a shared backbone. The network is a multi-layer perceptron~(MLP) consisting of hidden layers with 256, 128, and 64 units, each followed by ELU~\cite{clevert2015fast} activation. Currently, it takes about 5 minutes to train per task, which can be prohibitive for certain applications. However, this training time can be reduced by increasing the number of parallel environments and utilizing more powerful GPUs.

\textbf{Domain Randomization (DR)}: Recognizing the challenges inherent in transferring policies between the simulation and the real world, we employ DR to bridge the real-to-sim-to-real gaps. DR is applied to object properties like friction, mass, restitution, compliance, and geometry. We further randomize the object position, the gripper location, and the grasp within a range. We found these to be especially crucial for non-prehensile tasks like pushing. The specific parameter ranges are detailed in Appendix~\ref{sec:domain_randomization}, and the effectiveness of DR is evaluated in Section~\ref{sec:dr}.

\begin{table*}[h]
    \centering
    \begin{tabular}{lcccccccc}
        \toprule
        \multirow{2}{*}{\textbf{Task}} & \multicolumn{4}{c}{\textbf{Annotated (Human labeled reward)}} & \multicolumn{4}{c}{\textbf{Automatic (VLM-generated reward)}} \\
        \cmidrule(lr){2-5} \cmidrule(lr){6-9}
         & \multicolumn{2}{c}{\textbf{Simulation}} & \multicolumn{2}{c}{\textbf{Real-World}} & \multicolumn{2}{c}{\textbf{Simulation}} & \multicolumn{2}{c}{\textbf{Real-World}} \\
        \cmidrule(lr){2-3} \cmidrule(lr){4-5} \cmidrule(lr){6-7} \cmidrule(lr){8-9}
         & \textbf{\algabrvname (Ours)} & \textbf{Pose} & \textbf{\algabrvname (Ours)} & \textbf{Pose} & \textbf{\algabrvname (Ours)} & \textbf{Pose} & \textbf{\algabrvname (Ours)} & \textbf{Pose} \\
        \midrule
        \textbf{Shoe Place} & 0.945 & 0.938 & 0.8 & 0.9 & \textbf{0.778} & 0.353 & \textbf{0.7} & 0.3 \\
        \textbf{Shoe Push} & 0.871 & 0.850 & 0.7 & 0.7 & \textbf{0.716} & 0.289 & \textbf{0.6} & 0.2 \\
        \textbf{Stowing Push} & 0.901 & 0.914 & 0.8 & 0.7 & \textbf{0.679} & 0.374 & \textbf{0.6} & 0.3 \\
        \textbf{Stowing Reorient} & 0.848 & 0.859 & 0.8 & 0.7 & \textbf{0.858} & 0.265 & \textbf{0.7} & 0.2 \\
        \bottomrule
    \end{tabular}
    \caption{\small{\textbf{Performance of \algabrvname in simulation and real-world}. \algabrvname, which makes use of visual keypoints, significantly outperforms the conventional pose-based approach, especially when using VLMs to automatically generate reward functions.}}
    \label{tab:main_results}
    \vspace{-2em}
\end{table*}

\subsection{Deployment of Trained Policy}
\label{sec:deployment}
The trained RL policy $\pi_i$ is deployed directly in the real world.
Since the policy outputs the end-effector pose,
we employ inverse kinematics to compute the joint angles at each timestep. The RL policy operates at 10Hz, producing action commands that are then clipped to ensure the end effector remains within the workspace limits. For keypoint tracking, we utilize FoundationPose~\cite{wen2023foundationpose} to estimate the object’s pose. These pose estimates are subsequently used to compute the keypoint locations that are defined relative to the objects. When VLM predicts to grasp objects, we use AnyGrasp~\cite{fang2023anygrasp} to detect grasps in the real-world.

\section{Experiments and Analysis}
We aim to investigate whether \algname can effectively represent reward functions for diverse manipulation skills within our \algabrvname for real-to-sim-to-real pipeline. We also want to see whether our pipeline can  perform multi-step tasks in dynamic environments by leveraging \algname as feedback for replanning.

\subsection{Experimental Setup, Metrics and Baselines}
\begin{figure}[h]
    \centering
\includegraphics[width=0.75\linewidth]{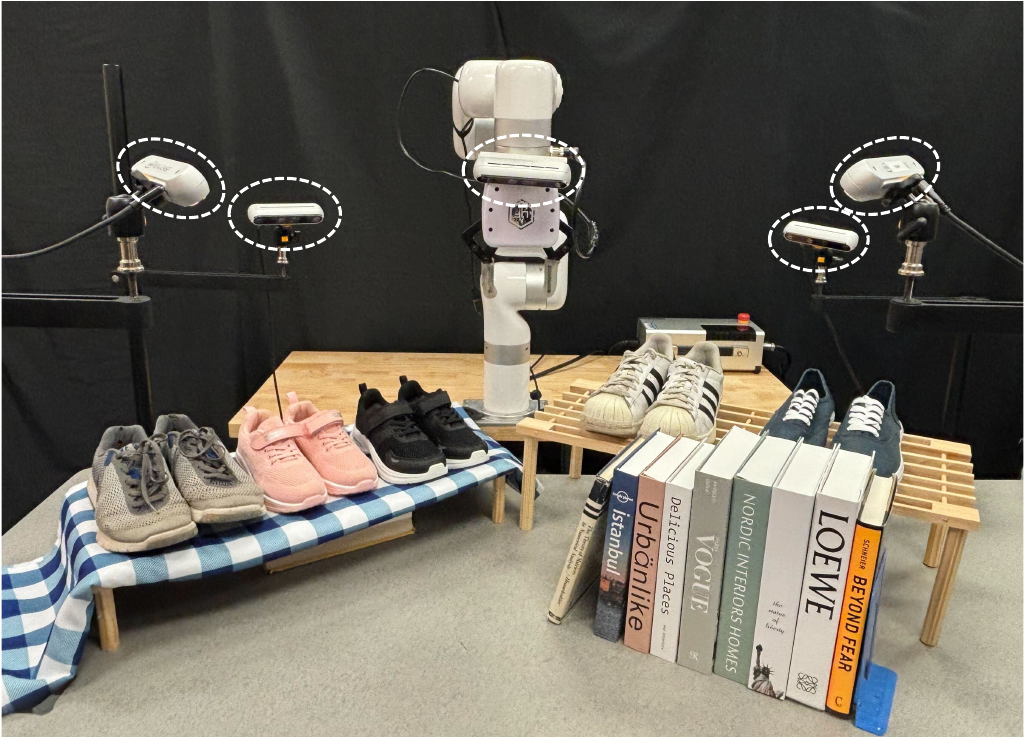}
    \caption{\small{\textbf{Setup and experiment objects.} We use XArm7 to conduct all our experiments. Our setup includes 4 stationary and 1 wrist-mounted camera. We experiment with 5 shoe pairs and 2 shoe racks for tasks involving shoe scenarios. Additionally, we experiment with 9 different books for stowing tasks.}}
    \label{fig:objects}
    \vspace{-0.8em}
\end{figure}

We conduct experiments on XArm7 with four stationary RealSense cameras. \figref{fig:objects} shows the setup, along with the objects used. These cameras capture the point clouds, which are used to construct the simulation environment and to provide data for AnyGrasp to predict grasp. Additionally, a wrist-mounted camera is used to capture images that are used to query the VLM. 

As a baseline, we use an \textit{annotated} variant of \algabrvname with human-labeled reward functions, allowing evaluation without VLM influence. We also compare our keypoint-based method with another baseline that uses object pose to construct reward function, which is more conventional in RL training~\cite{gu2017deep,popov2017data,vecerik2017leveraging,rajeswaran2017learning,qi2023hand}. In this pose-based method, the VLM generates a function $f$ that maps the initial object poses (represented by xyz coordinates for position and RPY angles for orientation) to their final poses. The prompt for this baseline is discussed in Sec.~\ref{sec:prompt}.

We evaluate our approach across four scenarios, illustrated in \figref{fig:teaser} (left): Shoe Place, Shoe Push, Book Push, and Book Reorient. In Shoe Place, the robot picks up a shoe from the ground and places it on a rack. In Shoe Push, it pushes a shoe towards other shoe to form a matching pair. In Book Push, it pushes a book to align with other book, or push the book towards table edge, and in Book Reorient, it repositions a book on a shelf. Each scenario has 10 start/end configurations. In simulation, success rates are averaged over 128 randomized environments generated for 10 start/end configuration, making a total of 1280 trials per scenario. In the real-world, success is evaluated directly on the 10 start/end configurations. A trial is considered successful in both cases if the average keypoint distance to the target is within 5 cm.

\subsection{Policy Training with \algabrvname for Single-Step Tasks}
We conduct experiments comparing RL training with keypoints and object pose in reward functions. Our experiments span four representative tasks, and are summarized in Table~\ref{tab:main_results}.

In the \textit{annotated} method, success rates for shoe placement using \algabrvname and object pose are 0.945 and 0.938, respectively. A similar trend is observed in the shoe push, stowing push, and reorient tasks, where performance differences are minimal. These results demonstrate that, when targets are specified through human annotations, both keypoints and object poses effectively capture the target locations and serve as viable approaches for RL policy training.

In the \textit{automatic} method, \algabrvname significantly outperforms object pose representations. For example, in shoe placement, \algabrvname achieves a 0.7 success rate, while object poses reach only 0.3. Similar results are seen across other tasks. Object pose success is limited to simpler scenarios with no orientation changes, as VLMs struggle with rotations in SO(3) space. In contrast, keypoints simplify the challenge by requiring VLMs to reason only in Cartesian space, eliminating the need to handle object poses in SE(3) space.

As shown in Table~\ref{tab:main_results}, there is a slight reduction in success rate from simulation to the real world. For shoe placement, \algabrvname achieves success of 0.945 in simulation and 0.8 in the real world. For shoe push, the success rate drops from 0.871 to 0.850. These results suggest that domain randomization described in Section~\ref{sec:training} helps the model generalize to real-world conditions, but factors like inaccuracies in environment reconstruction, real-world perception errors, and the inability to simulate extreme object dynamics still affect performance.

Most of the failures in our framework stem from discrepancies between the heuristic grasps used in simulation and the grasps generated by AnyGrasp in the real world, as well as incorrect VLM predictions. For incorrect VLM predictions, the model sometimes selects the wrong keypoints or fails to use all available keypoints on an object when determining its relationship to another object. For instance, if an object has four keypoints, the VLM may only use one of them, leading to suboptimal alignment and placement. These issues can be mitigated by providing more in-context examples while querying the VLMs.
These challenges may become less pronounced with the incorporation of advancements such as~\cite{chen2024spatialvlm}, which enhance the spatial reasoning capabilities of VLMs. Additionally, some failures are caused by physical dynamics when pushing objects. These issues can be partially mitigated by explicitly estimating dynamic parameters during real-to-sim transfer.

\subsection{Iterative Replanning for Multi-Step Tasks}

\begin{figure*}[t]
    \centering
    \includegraphics[width=\linewidth]{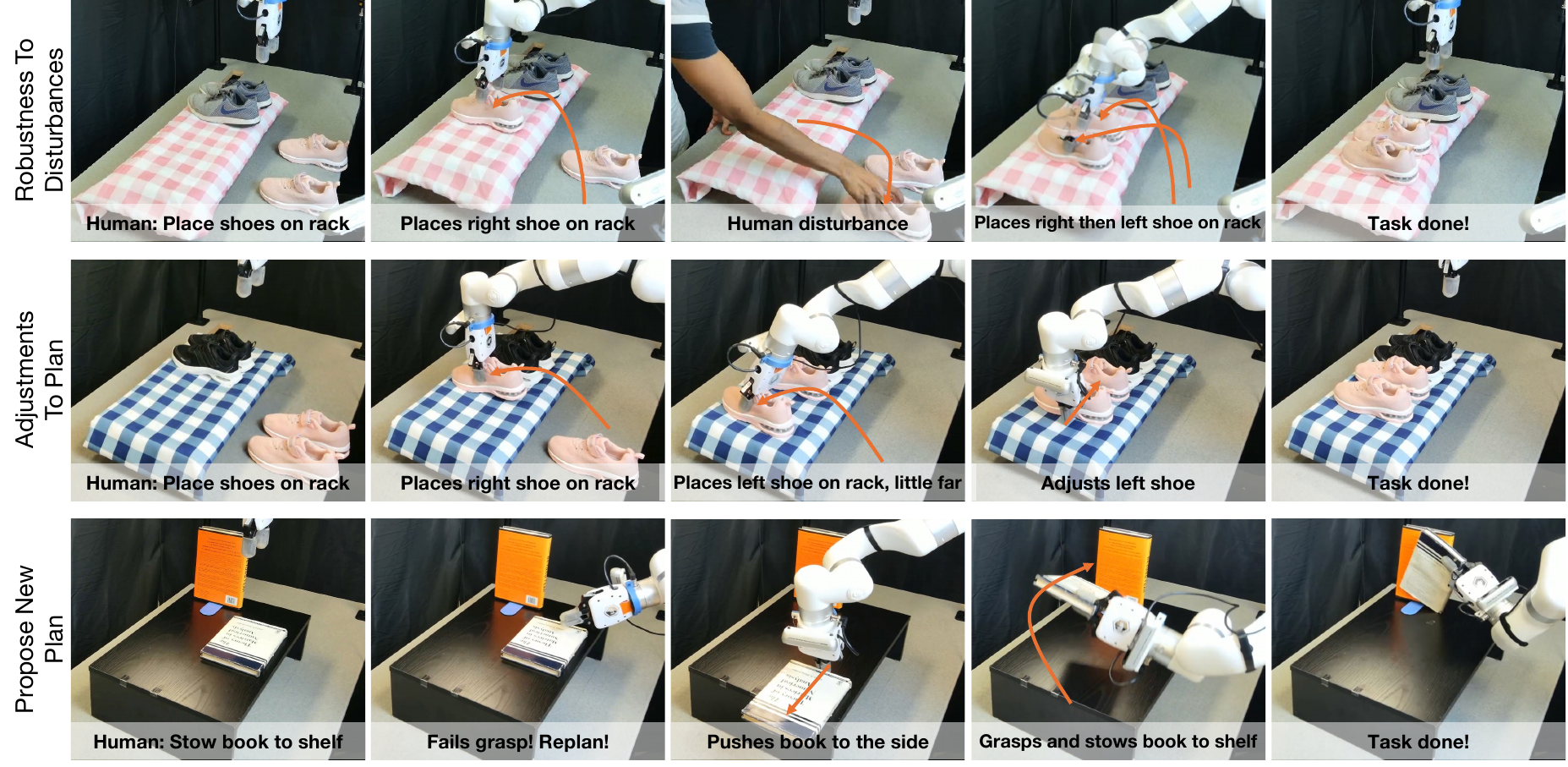}
    \caption{\small{\textbf{Scenarios demonstrating capabilities of our framework.} The framework is robust to disturbances and can adapt in response to unexpected events. Additionally, it can propose new plans when the original ones become infeasible.}}
    \label{fig:extensions}
    \vspace{-1.6em}
\end{figure*}

\begin{figure}[h]
    \setlength{\abovecaptionskip}{0pt}
    \includegraphics[width=\linewidth]{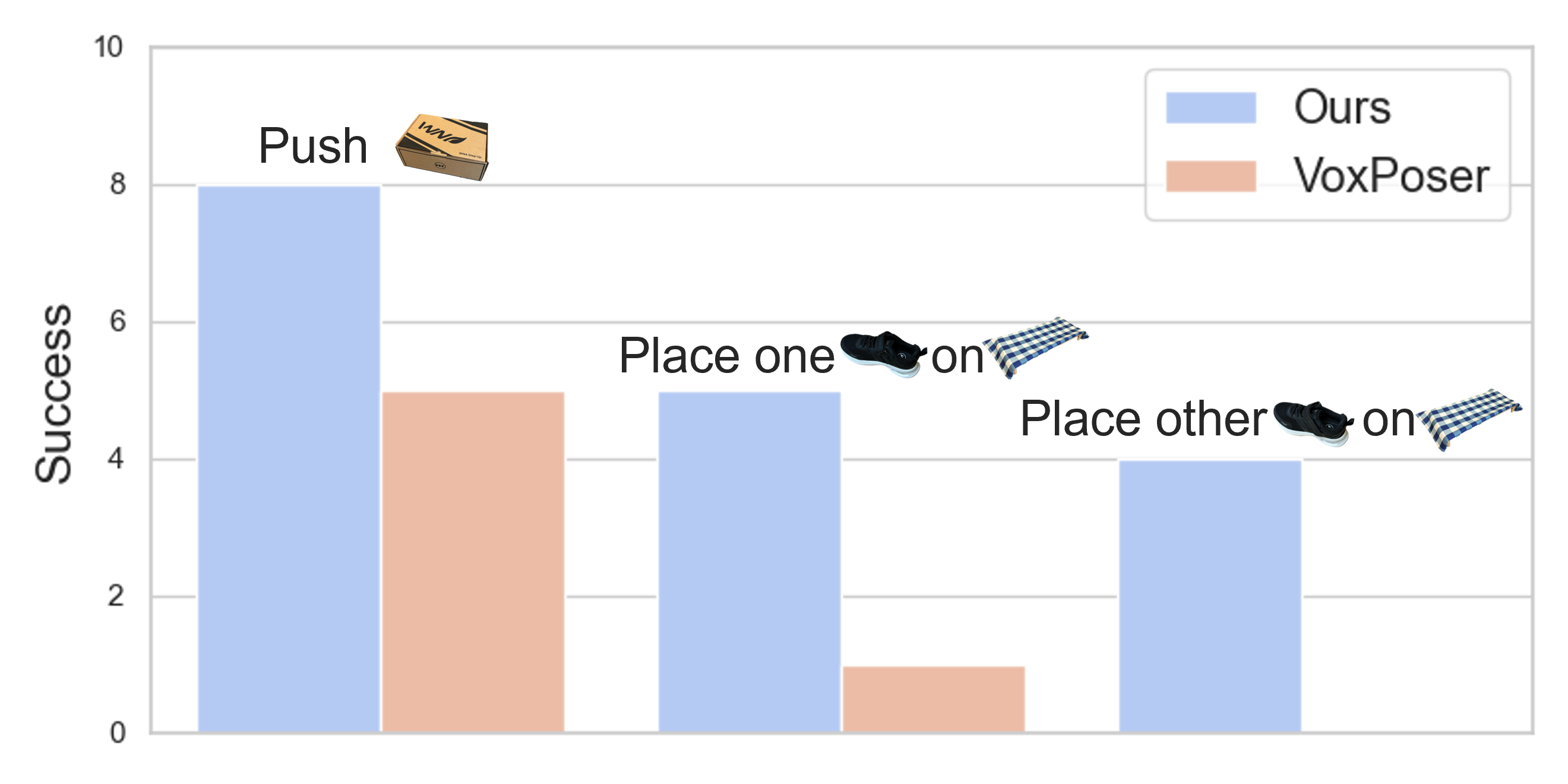}
    \caption{\small{\textbf{Multi-Step Task Chaining Comparison with VoxPoser.} Our proposed framework consistently demonstrates superior performance compared to VoxPoser at every step of the task sequence.}}
    \label{fig:chaining_analysis}
    \vspace{-2.1em}
\end{figure}

We demonstrate the robot's iterative chaining ability with a task of three sequential actions: first pushing a shoe box to create space, then placing a pair of shoes on a rack. Failure in one task leads to failure in the next. We evaluate this process using 10 different start and end configurations, iterating through each to assess overall performance.

We compare our method with VoxPoser~\cite{huang2023voxposer}, which employs LLMs to generate code that produces potential fields for motion planning. VoxPoser serves as an ideal baseline because it synthesizes motion plans for diverse manipulation tasks from free-form language instructions. Notably, VoxPoser plans are open-loop and lack feedback to refine specifications at each step. To adapt it to our tasks, we enhanced VoxPoser with two major modifications: (1) VoxPoser used OWL-ViT~\cite{minderer2022simple} to find object bounding boxes, but it struggled to distinguish between left and right shoes, so we provided ground-truth object locations. (2) We gave VoxPoser the entire plan, as the original planner struggled with multi-step tasks. This gave VoxPoser an advantage over our method due to access to privileged information.

\figref{fig:chaining_analysis} shows the iterative chaining results. Across the three tasks, our method consistently outperformed VoxPoser. In the first task, we succeeded 8 out of 10 times compared to VoxPoser's 5 successes. For the second task, we had 5 successes while VoxPoser had 1. In the final task, our method succeeded 4 times, whereas VoxPoser failed in all attempts. VoxPoser's failures can be attributed to several factors, such as pushing the shoe box either too far or not far enough. Additionally, its grasping strategy relies on a simple heuristic that positions the robot's end effector around the object center before closing the gripper, often resulting in failed grasps. It also struggles with collisions during object manipulation, as it does not account for the environment to avoid obstacles. Furthermore, improper placement of shoes—such as stacking both shoes on top of each other, causing them to fall—further highlights its limitations.

\subsection{Robustness, Adjusting Plans, and Re-Planning}
Unlike previous works that rely on open-loop plans, our approach leverages closed-loop plans, enabling adjustments during execution. This feature gives rise to several capabilities, as demonstrated in \figref{fig:extensions}.

In the first scenario, a human interrupts the robot while it is in the process of placing shoes on the ground. The framework demonstrates resilience by recovering from the interruption. The robot re-grasps the shoe and successfully completes the task by placing both shoes on the rack.

In the second scenario, when the robot attempts to place the left shoe, it detects that the shoe is not positioned close enough to the right shoe. To address this, the VLM predicts a corrective action, suggesting that the robot push the left shoe closer to the right shoe to form a proper pair.

In the third scenario, the robot is tasked with stowing a book on a shelf. However, the initial grasp attempt fails because the book is too large to be grasped. In response, the VLM predicts an alternative strategy to complete the task, adjusting the approach to ensure successful placement.

\subsection{Effect of Domain Randomization} \label{sec:dr}
\begin{table}[ht]
    \centering
    \begin{tabular}{lcccc}
        \toprule
        \multirow{2}{*}{\textbf{Task}} & \multicolumn{2}{c}{\textbf{Without DR}} & \multicolumn{2}{c}{\textbf{With DR}} \\
        \cmidrule(lr){2-3} \cmidrule(lr){4-5}
         & \textbf{Simulation} & \textbf{Real-World} & \textbf{Simulation} & \textbf{Real-World} \\
        \midrule
        \textbf{Place} & 0.964 & 0.5 & 0.945 & 0.8 \\
        \textbf{Push} & 0.923 & 0.2 & 0.871 & 0.7 \\
        \bottomrule
    \end{tabular}
    \caption{\small{\textbf{Performance of Shoe Place and Push with and without Domain Randomization (DR).} DR slightly reduces simulation performance but significantly improves real-world task performance across different scenarios.}}
    \label{tab:domain_randomization_results}
    \vspace{-2.5em}
\end{table}

We present the results of our framework with and without DR for shoe place and push. The performance is averaged over 10 runs. In the simulation, the performance without DR is 0.964, while with DR, it is slightly lower at 0.945, suggesting that without DR, the policy performs better in a single, controlled setting. However, in the real world, the performance without DR drops to 0.6, whereas with DR, it is 0.8, highlighting the effectiveness of DR. For the place task without DR, we observe that the policy is less robust to sim-to-real gap, frequently colliding with the shoe rack during transport. Additionally, immediately after picking up the object, the policy sometimes fails, likely due to discrepancies in the pose estimation.

For push, these issues are more pronounced: success is 0.2 without DR but improves to 0.7 with DR. Without DR, the policy often crushes the shoe or causes it to slip out of alignment during pushing. These findings demonstrate the importance of DR for reliable real-world performance.

\section{Conclusion and Limitations}
In this work, we introduced Iterative Keypoint Reward (\algabrvname), a framework that leverages VLMs to generate visually grounded reward functions for robotic manipulation in open-world environments. By using keypoints from RGB-D observations, our approach enables precise SE(3) control and integrates priors from VLMs without relying on rigid instructions. \algabrvname bridges simulation and real-world execution through a real-to-sim-to-real loop, training policies in simulation and deploying them in physical environments. Experiments across diverse tasks demonstrate the framework's ability to handle complex, long-horizon challenges with adaptive strategies and error recovery. This work represents a step toward more intelligent and flexible robots capable of operating effectively in dynamic, real-world settings.

Despite these advancements, our approach has certain limitations. We need to capture objects from all views to obtain object meshes. In the future, this may be simplified by using methods~\cite{liu2023zero} that can generate meshes from a single image. Additionally, our real-to-sim transfer does not account for dynamics parameters, which could be modeled more accurately through system identification techniques. Also, while our framework reconstructs multiple objects in the environment, we do not account for tasks involving complicated multi-object interactions, limiting our evaluation primarily to single-object manipulation at each stage. 

\section{Acknowledgements}
We thank Aditya Prakash, Arjun Gupta, Binghao Huang, Hanxiao Jiang, Kaifeng Zhang, and Unnat Jain for fruitful discussions. This work is partially supported by the Amazon AICE Award, the Sony Group Corporation, and the DARPA TIAMAT program (HR0011-24-9-0430). This work does not relate to the positions of Shubham Garg and Hooshang Nayyeri at Amazon. This article solely reflects the opinions and conclusions of its authors and should not be interpreted as necessarily representing the official policies, either expressed or implied, of the sponsors.

\bibliographystyle{IEEEtran}
\bibliography{IEEEabrv,references}

\begin{thebibliography}{100}
\providecommand{\url}[1]{#1}
\csname url@samestyle\endcsname
\providecommand{\newblock}{\relax}
\providecommand{\bibinfo}[2]{#2}
\providecommand{\BIBentrySTDinterwordspacing}{\spaceskip=0pt\relax}
\providecommand{\BIBentryALTinterwordstretchfactor}{4}
\providecommand{\BIBentryALTinterwordspacing}{\spaceskip=\fontdimen2\font plus
\BIBentryALTinterwordstretchfactor\fontdimen3\font minus \fontdimen4\font\relax}
\providecommand{\BIBforeignlanguage}[2]{{%
\expandafter\ifx\csname l@#1\endcsname\relax
\typeout{** WARNING: IEEEtran.bst: No hyphenation pattern has been}%
\typeout{** loaded for the language `#1'. Using the pattern for}%
\typeout{** the default language instead.}%
\else
\language=\csname l@#1\endcsname
\fi
#2}}
\providecommand{\BIBdecl}{\relax}
\BIBdecl

\bibitem{openai2023gpt}
OpenAI, ``Gpt-4 technical report,'' \emph{arXiv}, 2023.

\bibitem{zeng2022socratic}
A.~Zeng, A.~Wong, S.~Welker, K.~Choromanski, F.~Tombari, A.~Purohit, M.~Ryoo, V.~Sindhwani, J.~Lee, V.~Vanhoucke \emph{et~al.}, ``Socratic models: Composing zero-shot multimodal reasoning with language,'' \emph{arXiv preprint arXiv:2204.00598}, 2022.

\bibitem{radford2021learningtransferablevisualmodels}
\BIBentryALTinterwordspacing
A.~Radford, J.~W. Kim, C.~Hallacy, A.~Ramesh, G.~Goh, S.~Agarwal, G.~Sastry, A.~Askell, P.~Mishkin, J.~Clark, G.~Krueger, and I.~Sutskever, ``Learning transferable visual models from natural language supervision,'' 2021. [Online]. Available: \url{https://arxiv.org/abs/2103.00020}
\BIBentrySTDinterwordspacing

\bibitem{jia2021scalingvisualvisionlanguagerepresentation}
\BIBentryALTinterwordspacing
C.~Jia, Y.~Yang, Y.~Xia, Y.-T. Chen, Z.~Parekh, H.~Pham, Q.~V. Le, Y.~Sung, Z.~Li, and T.~Duerig, ``Scaling up visual and vision-language representation learning with noisy text supervision,'' 2021. [Online]. Available: \url{https://arxiv.org/abs/2102.05918}
\BIBentrySTDinterwordspacing

\bibitem{li2022blip}
J.~Li, D.~Li, C.~Xiong, and S.~Hoi, ``Blip: Bootstrapping language-image pre-training for unified vision-language understanding and generation,'' in \emph{International conference on machine learning}.\hskip 1em plus 0.5em minus 0.4em\relax PMLR, 2022, pp. 12\,888--12\,900.

\bibitem{li2023blip2bootstrappinglanguageimagepretraining}
\BIBentryALTinterwordspacing
J.~Li, D.~Li, S.~Savarese, and S.~Hoi, ``Blip-2: Bootstrapping language-image pre-training with frozen image encoders and large language models,'' 2023. [Online]. Available: \url{https://arxiv.org/abs/2301.12597}
\BIBentrySTDinterwordspacing

\bibitem{alayrac2022flamingovisuallanguagemodel}
\BIBentryALTinterwordspacing
J.-B. Alayrac, J.~Donahue, P.~Luc, A.~Miech, I.~Barr, Y.~Hasson, K.~Lenc, A.~Mensch, K.~Millican, M.~Reynolds, R.~Ring, E.~Rutherford, S.~Cabi, T.~Han, Z.~Gong, S.~Samangooei, M.~Monteiro, J.~Menick, S.~Borgeaud, A.~Brock, A.~Nematzadeh, S.~Sharifzadeh, M.~Binkowski, R.~Barreira, O.~Vinyals, A.~Zisserman, and K.~Simonyan, ``Flamingo: a visual language model for few-shot learning,'' 2022. [Online]. Available: \url{https://arxiv.org/abs/2204.14198}
\BIBentrySTDinterwordspacing

\bibitem{yu2022cocacontrastivecaptionersimagetext}
\BIBentryALTinterwordspacing
J.~Yu, Z.~Wang, V.~Vasudevan, L.~Yeung, M.~Seyedhosseini, and Y.~Wu, ``Coca: Contrastive captioners are image-text foundation models,'' 2022. [Online]. Available: \url{https://arxiv.org/abs/2205.01917}
\BIBentrySTDinterwordspacing

\bibitem{huang2024rekep}
W.~Huang, C.~Wang, Y.~Li, R.~Zhang, and L.~Fei-Fei, ``Rekep: Spatio-temporal reasoning of relational keypoint constraints for robotic manipulation,'' \emph{arXiv preprint arXiv:2409.01652}, 2024.

\bibitem{ahn2022can}
M.~Ahn, A.~Brohan, N.~Brown, Y.~Chebotar, O.~Cortes, B.~David, C.~Finn, C.~Fu, K.~Gopalakrishnan, K.~Hausman \emph{et~al.}, ``Do as i can, not as i say: Grounding language in robotic affordances,'' \emph{arXiv preprint arXiv:2204.01691}, 2022.

\bibitem{codeaspolicies2022}
J.~Liang, W.~Huang, F.~Xia, P.~Xu, K.~Hausman, B.~Ichter, P.~Florence, and A.~Zeng, ``Code as policies: Language model programs for embodied control,'' in \emph{arXiv preprint arXiv:2209.07753}, 2022.

\bibitem{huang2023voxposer}
W.~Huang, C.~Wang, R.~Zhang, Y.~Li, J.~Wu, and L.~Fei-Fei, ``Voxposer: Composable 3d value maps for robotic manipulation with language models,'' \emph{arXiv preprint arXiv:2307.05973}, 2023.

\bibitem{brohan2022rt}
A.~Brohan, N.~Brown, J.~Carbajal, Y.~Chebotar, J.~Dabis, C.~Finn, K.~Gopalakrishnan, K.~Hausman, A.~Herzog, J.~Hsu \emph{et~al.}, ``Rt-1: Robotics transformer for real-world control at scale,'' \emph{arXiv preprint arXiv:2212.06817}, 2022.

\bibitem{brohan2023rt}
A.~Brohan, N.~Brown, J.~Carbajal, Y.~Chebotar, X.~Chen, K.~Choromanski, T.~Ding, D.~Driess, A.~Dubey, C.~Finn \emph{et~al.}, ``Rt-2: Vision-language-action models transfer web knowledge to robotic control,'' \emph{arXiv preprint arXiv:2307.15818}, 2023.

\bibitem{liu2024moka}
F.~Liu, K.~Fang, P.~Abbeel, and S.~Levine, ``Moka: Open-vocabulary robotic manipulation through mark-based visual prompting,'' \emph{arXiv preprint arXiv:2403.03174}, 2024.

\bibitem{huang2024copa}
H.~Huang, F.~Lin, Y.~Hu, S.~Wang, and Y.~Gao, ``Copa: General robotic manipulation through spatial constraints of parts with foundation models,'' \emph{arXiv preprint arXiv:2403.08248}, 2024.

\bibitem{team2024octo}
O.~M. Team, D.~Ghosh, H.~Walke, K.~Pertsch, K.~Black, O.~Mees, S.~Dasari, J.~Hejna, T.~Kreiman, C.~Xu \emph{et~al.}, ``Octo: An open-source generalist robot policy,'' \emph{arXiv preprint arXiv:2405.12213}, 2024.

\bibitem{huang2023instruct2act}
S.~Huang, Z.~Jiang, H.~Dong, Y.~Qiao, P.~Gao, and H.~Li, ``Instruct2act: Mapping multi-modality instructions to robotic actions with large language model,'' \emph{arXiv preprint arXiv:2305.11176}, 2023.

\bibitem{xu2023creative}
M.~Xu, P.~Huang, W.~Yu, S.~Liu, X.~Zhang, Y.~Niu, T.~Zhang, F.~Xia, J.~Tan, and D.~Zhao, ``Creative robot tool use with large language models,'' \emph{arXiv preprint arXiv:2310.13065}, 2023.

\bibitem{zhou2023generalizable}
H.~Zhou, M.~Ding, W.~Peng, M.~Tomizuka, L.~Shao, and C.~Gan, ``Generalizable long-horizon manipulations with large language models,'' \emph{arXiv preprint arXiv:2310.02264}, 2023.

\bibitem{nasiriany2024pivot}
S.~Nasiriany, F.~Xia, W.~Yu, T.~Xiao, J.~Liang, I.~Dasgupta, A.~Xie, D.~Driess, A.~Wahid, Z.~Xu \emph{et~al.}, ``Pivot: Iterative visual prompting elicits actionable knowledge for vlms,'' \emph{arXiv preprint arXiv:2402.07872}, 2024.

\bibitem{di2024keypoint}
N.~Di~Palo and E.~Johns, ``Keypoint action tokens enable in-context imitation learning in robotics,'' \emph{arXiv preprint arXiv:2403.19578}, 2024.

\bibitem{zeng2023large}
F.~Zeng, W.~Gan, Y.~Wang, N.~Liu, and P.~S. Yu, ``Large language models for robotics: A survey,'' \emph{arXiv preprint arXiv:2311.07226}, 2023.

\bibitem{zha2024distilling}
L.~Zha, Y.~Cui, L.-H. Lin, M.~Kwon, M.~G. Arenas, A.~Zeng, F.~Xia, and D.~Sadigh, ``Distilling and retrieving generalizable knowledge for robot manipulation via language corrections,'' in \emph{2024 IEEE International Conference on Robotics and Automation (ICRA)}.\hskip 1em plus 0.5em minus 0.4em\relax IEEE, 2024, pp. 15\,172--15\,179.

\bibitem{arenas2024prompt}
M.~G. Arenas, T.~Xiao, S.~Singh, V.~Jain, A.~Ren, Q.~Vuong, J.~Varley, A.~Herzog, I.~Leal, S.~Kirmani \emph{et~al.}, ``How to prompt your robot: A promptbook for manipulation skills with code as policies,'' in \emph{2024 IEEE International Conference on Robotics and Automation (ICRA)}.\hskip 1em plus 0.5em minus 0.4em\relax IEEE, 2024, pp. 4340--4348.

\bibitem{mahadevan2024generative}
K.~Mahadevan, J.~Chien, N.~Brown, Z.~Xu, C.~Parada, F.~Xia, A.~Zeng, L.~Takayama, and D.~Sadigh, ``Generative expressive robot behaviors using large language models,'' in \emph{Proceedings of the 2024 ACM/IEEE International Conference on Human-Robot Interaction}, 2024, pp. 482--491.

\bibitem{liang2024learning}
J.~Liang, F.~Xia, W.~Yu, A.~Zeng, M.~G. Arenas, M.~Attarian, M.~Bauza, M.~Bennice, A.~Bewley, A.~Dostmohamed \emph{et~al.}, ``Learning to learn faster from human feedback with language model predictive control,'' \emph{arXiv preprint arXiv:2402.11450}, 2024.

\bibitem{huang2024grounded}
W.~Huang, F.~Xia, D.~Shah, D.~Driess, A.~Zeng, Y.~Lu, P.~Florence, I.~Mordatch, S.~Levine, K.~Hausman \emph{et~al.}, ``Grounded decoding: Guiding text generation with grounded models for embodied agents,'' \emph{Advances in Neural Information Processing Systems}, vol.~36, 2024.

\bibitem{ren2023robots}
A.~Z. Ren, A.~Dixit, A.~Bodrova, S.~Singh, S.~Tu, N.~Brown, P.~Xu, L.~Takayama, F.~Xia, J.~Varley \emph{et~al.}, ``Robots that ask for help: Uncertainty alignment for large language model planners,'' \emph{arXiv preprint arXiv:2307.01928}, 2023.

\bibitem{jiang2022vima}
Y.~Jiang, A.~Gupta, Z.~Zhang, G.~Wang, Y.~Dou, Y.~Chen, L.~Fei-Fei, A.~Anandkumar, Y.~Zhu, and L.~Fan, ``Vima: General robot manipulation with multimodal prompts,'' \emph{arXiv preprint arXiv:2210.03094}, vol.~2, no.~3, p.~6, 2022.

\bibitem{yang2024guiding}
Z.~Yang, C.~Garrett, D.~Fox, T.~Lozano-P{\'e}rez, and L.~P. Kaelbling, ``Guiding long-horizon task and motion planning with vision language models,'' \emph{arXiv preprint arXiv:2410.02193}, 2024.

\bibitem{duan2024aha}
J.~Duan, W.~Pumacay, N.~Kumar, Y.~R. Wang, S.~Tian, W.~Yuan, R.~Krishna, D.~Fox, A.~Mandlekar, and Y.~Guo, ``Aha: A vision-language-model for detecting and reasoning over failures in robotic manipulation,'' \emph{arXiv preprint arXiv:2410.00371}, 2024.

\bibitem{duan2024manipulate}
J.~Duan, W.~Yuan, W.~Pumacay, Y.~R. Wang, K.~Ehsani, D.~Fox, and R.~Krishna, ``Manipulate-anything: Automating real-world robots using vision-language models,'' \emph{arXiv preprint arXiv:2406.18915}, 2024.

\bibitem{yuan2024robopoint}
W.~Yuan, J.~Duan, V.~Blukis, W.~Pumacay, R.~Krishna, A.~Murali, A.~Mousavian, and D.~Fox, ``Robopoint: A vision-language model for spatial affordance prediction for robotics,'' \emph{arXiv preprint arXiv:2406.10721}, 2024.

\bibitem{singh2023progprompt}
I.~Singh, V.~Blukis, A.~Mousavian, A.~Goyal, D.~Xu, J.~Tremblay, D.~Fox, J.~Thomason, and A.~Garg, ``Progprompt: program generation for situated robot task planning using large language models,'' \emph{Autonomous Robots}, vol.~47, no.~8, pp. 999--1012, 2023.

\bibitem{tang2024kalie}
G.~Tang, S.~Rajkumar, Y.~Zhou, H.~R. Walke, S.~Levine, and K.~Fang, ``Kalie: Fine-tuning vision-language models for open-world manipulation without robot data,'' \emph{arXiv preprint arXiv:2409.14066}, 2024.

\bibitem{liang2024eurekaverse}
W.~Liang, S.~Wang, H.-J. Wang, O.~Bastani, D.~Jayaraman, and Y.~J. Ma, ``Eurekaverse: Environment curriculum generation via large language models,'' \emph{arXiv preprint arXiv:2411.01775}, 2024.

\bibitem{zawalski2024robotic}
M.~Zawalski, W.~Chen, K.~Pertsch, O.~Mees, C.~Finn, and S.~Levine, ``Robotic control via embodied chain-of-thought reasoning,'' \emph{arXiv preprint arXiv:2407.08693}, 2024.

\bibitem{o2023open}
A.~O'Neill, A.~Rehman, A.~Gupta, A.~Maddukuri, A.~Gupta, A.~Padalkar, A.~Lee, A.~Pooley, A.~Gupta, A.~Mandlekar \emph{et~al.}, ``Open x-embodiment: Robotic learning datasets and rt-x models,'' \emph{arXiv preprint arXiv:2310.08864}, 2023.

\bibitem{huang2022language}
W.~Huang, P.~Abbeel, D.~Pathak, and I.~Mordatch, ``Language models as zero-shot planners: Extracting actionable knowledge for embodied agents,'' in \emph{International Conference on Machine Learning}.\hskip 1em plus 0.5em minus 0.4em\relax PMLR, 2022.

\bibitem{rth2024arxiv}
S.~Belkhale, T.~Ding, T.~Xiao, P.~Sermanet, Q.~Vuong, J.~Tompson, Y.~Chebotar, D.~Dwibedi, and D.~Sadigh, ``Rt-h: Action hierarchies using language,'' in \emph{https://arxiv.org/abs/2403.01823}, 2024.

\bibitem{zhao2024vlmpc}
W.~Zhao, J.~Chen, Z.~Meng, D.~Mao, R.~Song, and W.~Zhang, ``Vlmpc: Vision-language model predictive control for robotic manipulation,'' in \emph{Robotics: Science and Systems}, 2024.

\bibitem{yu2023language}
W.~Yu, N.~Gileadi, C.~Fu, S.~Kirmani, K.-H. Lee, M.~G. Arenas, H.-T.~L. Chiang, T.~Erez, L.~Hasenclever, J.~Humplik \emph{et~al.}, ``Language to rewards for robotic skill synthesis,'' \emph{arXiv preprint arXiv:2306.08647}, 2023.

\bibitem{ma2023eureka}
Y.~J. Ma, W.~Liang, G.~Wang, D.-A. Huang, O.~Bastani, D.~Jayaraman, Y.~Zhu, L.~Fan, and A.~Anandkumar, ``Eureka: Human-level reward design via coding large language models,'' \emph{arXiv preprint arXiv: Arxiv-2310.12931}, 2023.

\bibitem{ma2024dreurekalanguagemodelguided}
\BIBentryALTinterwordspacing
Y.~J. Ma, W.~Liang, H.-J. Wang, S.~Wang, Y.~Zhu, L.~Fan, O.~Bastani, and D.~Jayaraman, ``Dreureka: Language model guided sim-to-real transfer,'' 2024. [Online]. Available: \url{https://arxiv.org/abs/2406.01967}
\BIBentrySTDinterwordspacing

\bibitem{xie2023text2reward}
T.~Xie, S.~Zhao, C.~H. Wu, Y.~Liu, Q.~Luo, V.~Zhong, Y.~Yang, and T.~Yu, ``Text2reward: Automated dense reward function generation for reinforcement learning,'' \emph{arXiv preprint arXiv:2309.11489}, 2023.

\bibitem{kappler2018real}
D.~Kappler, F.~Meier, J.~Issac, J.~Mainprice, C.~G. Cifuentes, M.~W{\"u}thrich, V.~Berenz, S.~Schaal, N.~Ratliff, and J.~Bohg, ``Real-time perception meets reactive motion generation,'' \emph{IEEE Robotics and Automation Letters}, vol.~3, no.~3, pp. 1864--1871, 2018.

\bibitem{wen2022you}
B.~Wen, W.~Lian, K.~Bekris, and S.~Schaal, ``You only demonstrate once: Category-level manipulation from single visual demonstration,'' \emph{arXiv preprint arXiv:2201.12716}, 2022.

\bibitem{liu2024one}
M.~Liu, R.~Shi, L.~Chen, Z.~Zhang, C.~Xu, X.~Wei, H.~Chen, C.~Zeng, J.~Gu, and H.~Su, ``One-2-3-45++: Fast single image to 3d objects with consistent multi-view generation and 3d diffusion,'' in \emph{Proceedings of the IEEE/CVF Conference on Computer Vision and Pattern Recognition}, 2024, pp. 10\,072--10\,083.

\bibitem{xu2025sparp}
C.~Xu, A.~Li, L.~Chen, Y.~Liu, R.~Shi, H.~Su, and M.~Liu, ``Sparp: Fast 3d object reconstruction and pose estimation from sparse views,'' in \emph{European Conference on Computer Vision}.\hskip 1em plus 0.5em minus 0.4em\relax Springer, 2025, pp. 143--163.

\bibitem{shi2023zero123++}
R.~Shi, H.~Chen, Z.~Zhang, M.~Liu, C.~Xu, X.~Wei, L.~Chen, C.~Zeng, and H.~Su, ``Zero123++: a single image to consistent multi-view diffusion base model,'' \emph{arXiv preprint arXiv:2310.15110}, 2023.

\bibitem{liu2023zero}
R.~Liu, R.~Wu, B.~Van~Hoorick, P.~Tokmakov, S.~Zakharov, and C.~Vondrick, ``Zero-1-to-3: Zero-shot one image to 3d object,'' in \emph{Proceedings of the IEEE/CVF international conference on computer vision}, 2023, pp. 9298--9309.

\bibitem{gao2022get3d}
J.~Gao, T.~Shen, Z.~Wang, W.~Chen, K.~Yin, D.~Li, O.~Litany, Z.~Gojcic, and S.~Fidler, ``Get3d: A generative model of high quality 3d textured shapes learned from images,'' \emph{Advances In Neural Information Processing Systems}, vol.~35, pp. 31\,841--31\,854, 2022.

\bibitem{mu2021sdf}
J.~Mu, W.~Qiu, A.~Kortylewski, A.~Yuille, N.~Vasconcelos, and X.~Wang, ``A-sdf: Learning disentangled signed distance functions for articulated shape representation,'' in \emph{Proceedings of the IEEE/CVF International Conference on Computer Vision}, 2021, pp. 13\,001--13\,011.

\bibitem{jiang2022ditto}
Z.~Jiang, C.-C. Hsu, and Y.~Zhu, ``Ditto: Building digital twins of articulated objects from interaction,'' in \emph{Proceedings of the IEEE/CVF Conference on Computer Vision and Pattern Recognition}, 2022, pp. 5616--5626.

\bibitem{nie2022structure}
N.~Nie, S.~Y. Gadre, K.~Ehsani, and S.~Song, ``Structure from action: Learning interactions for articulated object 3d structure discovery,'' \emph{arXiv preprint arXiv:2207.08997}, 2022.

\bibitem{chen2024urdformer}
Z.~Chen, A.~Walsman, M.~Memmel, K.~Mo, A.~Fang, K.~Vemuri, A.~Wu, D.~Fox, and A.~Gupta, ``Urdformer: A pipeline for constructing articulated simulation environments from real-world images,'' \emph{arXiv preprint arXiv:2405.11656}, 2024.

\bibitem{mandi2024real2code}
Z.~Mandi, Y.~Weng, D.~Bauer, and S.~Song, ``Real2code: Reconstruct articulated objects via code generation,'' \emph{arXiv preprint arXiv:2406.08474}, 2024.

\bibitem{liu2023paris}
J.~Liu, A.~Mahdavi-Amiri, and M.~Savva, ``Paris: Part-level reconstruction and motion analysis for articulated objects,'' in \emph{Proceedings of the IEEE/CVF International Conference on Computer Vision}, 2023, pp. 352--363.

\bibitem{liu2024cage}
J.~Liu, H.~I.~I. Tam, A.~Mahdavi-Amiri, and M.~Savva, ``Cage: Controllable articulation generation,'' in \emph{Proceedings of the IEEE/CVF Conference on Computer Vision and Pattern Recognition}, 2024, pp. 17\,880--17\,889.

\bibitem{liu2024singapo}
J.~Liu, D.~Iliash, A.~X. Chang, M.~Savva, and A.~Mahdavi-Amiri, ``Singapo: Single image controlled generation of articulated parts in object,'' \emph{arXiv preprint arXiv:2410.16499}, 2024.

\bibitem{huang2012occlusion}
X.~Huang, I.~Walker, and S.~Birchfield, ``Occlusion-aware reconstruction and manipulation of 3d articulated objects,'' in \emph{2012 IEEE international conference on robotics and automation}.\hskip 1em plus 0.5em minus 0.4em\relax IEEE, 2012, pp. 1365--1371.

\bibitem{wen2023bundlesdf}
B.~Wen, J.~Tremblay, V.~Blukis, S.~Tyree, T.~Muller, A.~Evans, D.~Fox, J.~Kautz, and S.~Birchfield, ``Bundlesdf: Neural 6-dof tracking and 3d reconstruction of unknown objects,'' \emph{CVPR}, 2023.

\bibitem{shridhar2021cliport}
M.~Shridhar, L.~Manuelli, and D.~Fox, ``Cliport: What and where pathways for robotic manipulation,'' \emph{arXiv preprint arXiv: Arxiv-2109.12098}, 2021.

\bibitem{jiang2024transicsimtorealpolicytransfer}
\BIBentryALTinterwordspacing
Y.~Jiang, C.~Wang, R.~Zhang, J.~Wu, and L.~Fei-Fei, ``Transic: Sim-to-real policy transfer by learning from online correction,'' 2024. [Online]. Available: \url{https://arxiv.org/abs/2405.10315}
\BIBentrySTDinterwordspacing

\bibitem{gu2022multiskill}
J.~Gu, D.~S. Chaplot, H.~Su, and J.~Malik, ``Multi-skill mobile manipulation for object rearrangement,'' \emph{arXiv preprint arXiv: Arxiv-2209.02778}, 2022.

\bibitem{yenamandra2023homerobot}
S.~Yenamandra, A.~Ramachandran, K.~Yadav, A.~Wang, M.~Khanna, T.~Gervet, T.-Y. Yang, V.~Jain, A.~W. Clegg, J.~Turner, Z.~Kira, M.~Savva, A.~Chang, D.~S. Chaplot, D.~Batra, R.~Mottaghi, Y.~Bisk, and C.~Paxton, ``Homerobot: Open-vocabulary mobile manipulation,'' \emph{arXiv preprint arXiv: Arxiv-2306.11565}, 2023.

\bibitem{huang2023dynamic}
B.~Huang, Y.~Chen, T.~Wang, Y.~Qin, Y.~Yang, N.~Atanasov, and X.~Wang, ``Dynamic handover: Throw and catch with bimanual hands,'' \emph{arXiv preprint arXiv:2309.05655}, 2023.

\bibitem{chen2023sequential}
Y.~Chen, C.~Wang, L.~Fei-Fei, and C.~K. Liu, ``Sequential dexterity: Chaining dexterous policies for long-horizon manipulation,'' \emph{arXiv preprint arXiv:2309.00987}, 2023.

\bibitem{qin2022dexpointgeneralizablepointcloud}
\BIBentryALTinterwordspacing
Y.~Qin, B.~Huang, Z.-H. Yin, H.~Su, and X.~Wang, ``Dexpoint: Generalizable point cloud reinforcement learning for sim-to-real dexterous manipulation,'' 2022. [Online]. Available: \url{https://arxiv.org/abs/2211.09423}
\BIBentrySTDinterwordspacing

\bibitem{qi2023hand}
H.~Qi, A.~Kumar, R.~Calandra, Y.~Ma, and J.~Malik, ``In-hand object rotation via rapid motor adaptation,'' in \emph{Conference on Robot Learning}.\hskip 1em plus 0.5em minus 0.4em\relax PMLR, 2023, pp. 1722--1732.

\bibitem{yin2023rotating}
Z.-H. Yin, B.~Huang, Y.~Qin, Q.~Chen, and X.~Wang, ``Rotating without seeing: Towards in-hand dexterity through touch,'' \emph{arXiv preprint arXiv:2303.10880}, 2023.

\bibitem{DBLP:conf/rss/KumarFPM21}
\BIBentryALTinterwordspacing
A.~Kumar, Z.~Fu, D.~Pathak, and J.~Malik, ``{RMA:} rapid motor adaptation for legged robots,'' in \emph{Robotics: Science and Systems XVII, Virtual Event, July 12-16, 2021}, D.~A. Shell, M.~Toussaint, and M.~A. Hsieh, Eds., 2021. [Online]. Available: \url{https://doi.org/10.15607/RSS.2021.XVII.011}
\BIBentrySTDinterwordspacing

\bibitem{he2024agilesafelearningcollisionfree}
\BIBentryALTinterwordspacing
T.~He, C.~Zhang, W.~Xiao, G.~He, C.~Liu, and G.~Shi, ``Agile but safe: Learning collision-free high-speed legged locomotion,'' 2024. [Online]. Available: \url{https://arxiv.org/abs/2401.17583}
\BIBentrySTDinterwordspacing

\bibitem{tan2018simtoreal}
J.~Tan, T.~Zhang, E.~Coumans, A.~Iscen, Y.~Bai, D.~Hafner, S.~Bohez, and V.~Vanhoucke, ``Sim-to-real: Learning agile locomotion for quadruped robots,'' \emph{arXiv preprint arXiv: Arxiv-1804.10332}, 2018.

\bibitem{chang2020sim2real2sim}
P.~Chang and T.~Padir, ``Sim2real2sim: Bridging the gap between simulation and real-world in flexible object manipulation,'' \emph{arXiv preprint arXiv: Arxiv-2002.02538}, 2020.

\bibitem{lim2021planar}
V.~Lim, H.~Huang, L.~Y. Chen, J.~Wang, J.~Ichnowski, D.~Seita, M.~Laskey, and K.~Goldberg, ``Planar robot casting with real2sim2real self-supervised learning,'' \emph{arXiv preprint arXiv: Arxiv-2111.04814}, 2021.

\bibitem{bousmalis2018using}
K.~Bousmalis, A.~Irpan, P.~Wohlhart, Y.~Bai, M.~Kelcey, M.~Kalakrishnan, L.~Downs, J.~Ibarz, P.~Pastor, K.~Konolige \emph{et~al.}, ``Using simulation and domain adaptation to improve efficiency of deep robotic grasping,'' in \emph{2018 IEEE international conference on robotics and automation (ICRA)}.\hskip 1em plus 0.5em minus 0.4em\relax IEEE, 2018, pp. 4243--4250.

\bibitem{arndt2019metareinforcementlearningsimtoreal}
\BIBentryALTinterwordspacing
K.~Arndt, M.~Hazara, A.~Ghadirzadeh, and V.~Kyrki, ``Meta reinforcement learning for sim-to-real domain adaptation,'' 2019. [Online]. Available: \url{https://arxiv.org/abs/1909.12906}
\BIBentrySTDinterwordspacing

\bibitem{rao2020rlcycleganreinforcementlearningaware}
\BIBentryALTinterwordspacing
K.~Rao, C.~Harris, A.~Irpan, S.~Levine, J.~Ibarz, and M.~Khansari, ``Rl-cyclegan: Reinforcement learning aware simulation-to-real,'' 2020. [Online]. Available: \url{https://arxiv.org/abs/2006.09001}
\BIBentrySTDinterwordspacing

\bibitem{james2019sim}
S.~James, P.~Wohlhart, M.~Kalakrishnan, D.~Kalashnikov, A.~Irpan, J.~Ibarz, S.~Levine, R.~Hadsell, and K.~Bousmalis, ``Sim-to-real via sim-to-sim: Data-efficient robotic grasping via randomized-to-canonical adaptation networks,'' in \emph{Proceedings of the IEEE/CVF conference on computer vision and pattern recognition}, 2019, pp. 12\,627--12\,637.

\bibitem{du2022bayesian}
Y.~Du, D.~Ho, A.~Alemi, E.~Jang, and M.~Khansari, ``Bayesian imitation learning for end-to-end mobile manipulation,'' in \emph{International Conference on Machine Learning}.\hskip 1em plus 0.5em minus 0.4em\relax PMLR, 2022, pp. 5531--5546.

\bibitem{openai2019solving}
OpenAI, I.~Akkaya, M.~Andrychowicz, M.~Chociej, M.~Litwin, B.~McGrew, A.~Petron, A.~Paino, M.~Plappert, G.~Powell, R.~Ribas, J.~Schneider, N.~Tezak, J.~Tworek, P.~Welinder, L.~Weng, Q.~Yuan, W.~Zaremba, and L.~Zhang, ``Solving rubik's cube with a robot hand,'' \emph{arXiv preprint arXiv: Arxiv-1910.07113}, 2019.

\bibitem{tobin2017domain}
J.~Tobin, R.~Fong, A.~Ray, J.~Schneider, W.~Zaremba, and P.~Abbeel, ``Domain randomization for transferring deep neural networks from simulation to the real world,'' in \emph{2017 IEEE/RSJ international conference on intelligent robots and systems (IROS)}.\hskip 1em plus 0.5em minus 0.4em\relax IEEE, 2017, pp. 23--30.

\bibitem{antonova2021bayessimig}
R.~Antonova, F.~Ramos, R.~Possas, and D.~Fox, ``Bayessimig: Scalable parameter inference for adaptive domain randomization with isaacgym,'' \emph{arXiv preprint arXiv:2107.04527}, 2021.

\bibitem{lee2020learning}
J.~Lee, J.~Hwangbo, L.~Wellhausen, V.~Koltun, and M.~Hutter, ``Learning quadrupedal locomotion over challenging terrain,'' \emph{Science robotics}, vol.~5, no.~47, p. eabc5986, 2020.

\bibitem{peng2018sim}
X.~B. Peng, M.~Andrychowicz, W.~Zaremba, and P.~Abbeel, ``Sim-to-real transfer of robotic control with dynamics randomization,'' in \emph{2018 IEEE international conference on robotics and automation (ICRA)}.\hskip 1em plus 0.5em minus 0.4em\relax IEEE, 2018, pp. 3803--3810.

\bibitem{chebotar2019closing}
Y.~Chebotar, A.~Handa, V.~Makoviychuk, M.~Macklin, J.~Issac, N.~Ratliff, and D.~Fox, ``Closing the sim-to-real loop: Adapting simulation randomization with real world experience,'' in \emph{2019 International Conference on Robotics and Automation (ICRA)}.\hskip 1em plus 0.5em minus 0.4em\relax IEEE, 2019, pp. 8973--8979.

\bibitem{torne2024reconciling}
M.~Torne, A.~Simeonov, Z.~Li, A.~Chan, T.~Chen, A.~Gupta, and P.~Agrawal, ``Reconciling reality through simulation: A real-to-sim-to-real approach for robust manipulation,'' \emph{arXiv preprint arXiv:2403.03949}, 2024.

\bibitem{wen2023foundationpose}
B.~Wen, W.~Yang, J.~Kautz, and S.~Birchfield, ``Foundationpose: Unified 6d pose estimation and tracking of novel objects,'' \emph{arXiv preprint arXiv:2312.08344}, 2023.

\bibitem{makoviychuk2021isaac}
V.~Makoviychuk, L.~Wawrzyniak, Y.~Guo, M.~Lu, K.~Storey, M.~Macklin, D.~Hoeller, N.~Rudin, A.~Allshire, A.~Handa, and G.~State, ``Isaac gym: High performance gpu-based physics simulation for robot learning,'' 2021.

\bibitem{schulman2017proximal}
J.~Schulman, F.~Wolski, P.~Dhariwal, A.~Radford, and O.~Klimov, ``Proximal policy optimization algorithms,'' \emph{arXiv preprint arXiv:1707.06347}, 2017.

\bibitem{konda1999actor}
V.~Konda and J.~Tsitsiklis, ``Actor-critic algorithms,'' \emph{Advances in neural information processing systems}, vol.~12, 1999.

\bibitem{clevert2015fast}
D.-A. Clevert, ``Fast and accurate deep network learning by exponential linear units (elus),'' \emph{arXiv preprint arXiv:1511.07289}, 2015.

\bibitem{fang2023anygrasp}
H.-S. Fang, C.~Wang, H.~Fang, M.~Gou, J.~Liu, H.~Yan, W.~Liu, Y.~Xie, and C.~Lu, ``Anygrasp: Robust and efficient grasp perception in spatial and temporal domains,'' \emph{IEEE Transactions on Robotics}, 2023.

\bibitem{gu2017deep}
S.~Gu, E.~Holly, T.~Lillicrap, and S.~Levine, ``Deep reinforcement learning for robotic manipulation with asynchronous off-policy updates,'' in \emph{2017 IEEE international conference on robotics and automation (ICRA)}.\hskip 1em plus 0.5em minus 0.4em\relax IEEE, 2017, pp. 3389--3396.

\bibitem{popov2017data}
I.~Popov, N.~Heess, T.~Lillicrap, R.~Hafner, G.~Barth-Maron, M.~Vecerik, T.~Lampe, Y.~Tassa, T.~Erez, and M.~Riedmiller, ``Data-efficient deep reinforcement learning for dexterous manipulation,'' \emph{arXiv preprint arXiv:1704.03073}, 2017.

\bibitem{vecerik2017leveraging}
M.~Vecerik, T.~Hester, J.~Scholz, F.~Wang, O.~Pietquin, B.~Piot, N.~Heess, T.~Roth{\"o}rl, T.~Lampe, and M.~Riedmiller, ``Leveraging demonstrations for deep reinforcement learning on robotics problems with sparse rewards,'' \emph{arXiv preprint arXiv:1707.08817}, 2017.

\bibitem{rajeswaran2017learning}
A.~Rajeswaran, V.~Kumar, A.~Gupta, G.~Vezzani, J.~Schulman, E.~Todorov, and S.~Levine, ``Learning complex dexterous manipulation with deep reinforcement learning and demonstrations,'' \emph{arXiv preprint arXiv:1709.10087}, 2017.

\bibitem{chen2024spatialvlm}
B.~Chen, Z.~Xu, S.~Kirmani, B.~Ichter, D.~Sadigh, L.~Guibas, and F.~Xia, ``Spatialvlm: Endowing vision-language models with spatial reasoning capabilities,'' in \emph{Proceedings of the IEEE/CVF Conference on Computer Vision and Pattern Recognition}, 2024, pp. 14\,455--14\,465.

\bibitem{minderer2022simple}
M.~Minderer, A.~Gritsenko, A.~Stone, M.~Neumann, D.~Weissenborn, A.~Dosovitskiy, A.~Mahendran, A.~Arnab, M.~Dehghani, Z.~Shen \emph{et~al.}, ``Simple open-vocabulary object detection with vision transformers,'' \emph{arXiv preprint arXiv:2205.06230}, 2022.

\end{thebibliography}

\clearpage
\appendix

\begin{figure*}[h!]
    \setlength{\abovecaptionskip}{0pt}
    \includegraphics[width=\linewidth]{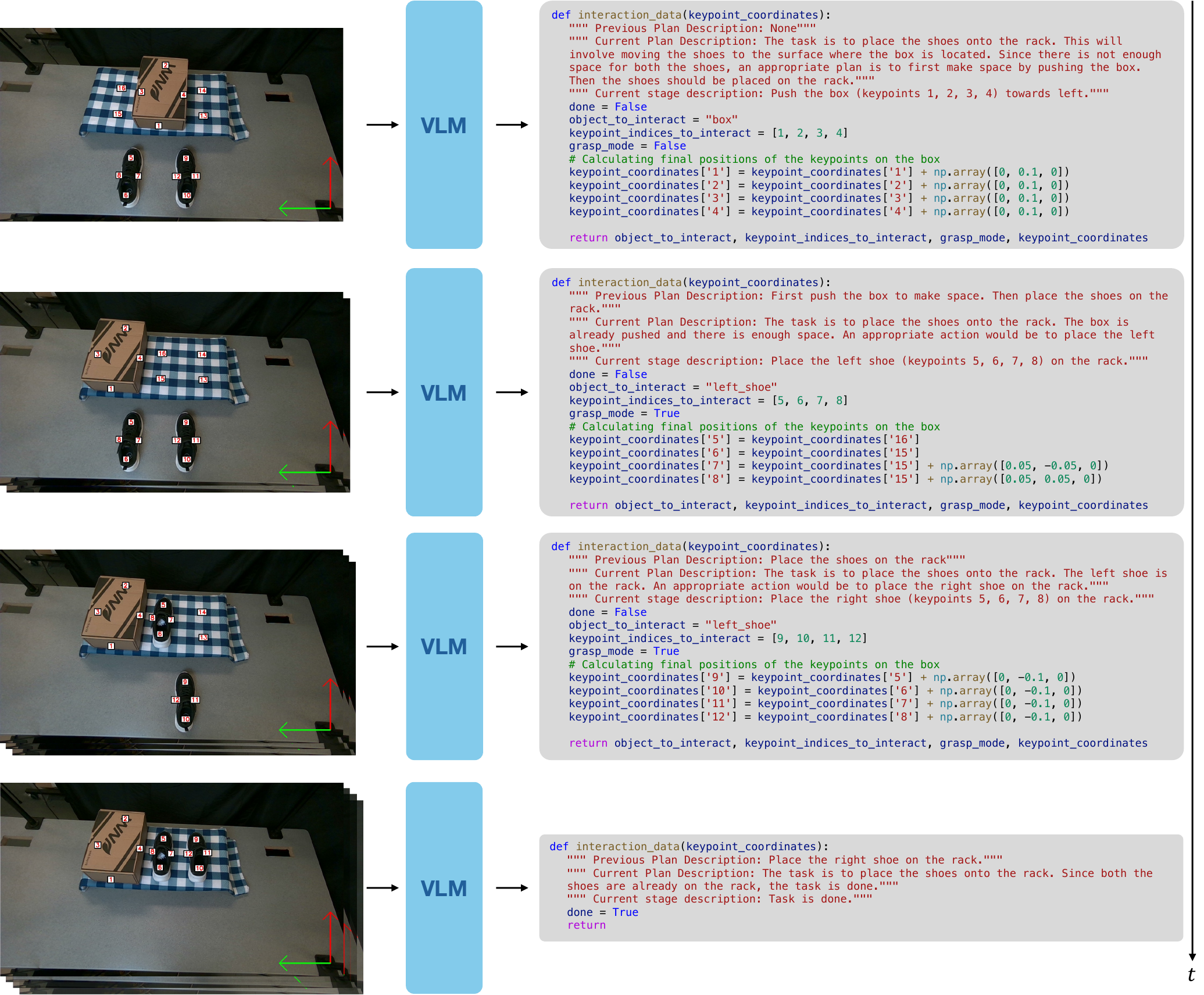}
    \vspace{0.1em}
    \caption{\small{\textbf{Examples of keypoint-marked images with corresponding predicted codes.} The top row represents the starting point, with subsequent rows illustrating the progression step by step.  The VLM first predicts to push the box to create space, followed by sequential placement of the shoes.}
    \label{fig:unrolled}}
    
\end{figure*}

\subsection{Grasping Subroutine}
\label{sec:grasping}

During training, the gripper fingers open only in the \textit{grasp} mode, where the end-effector approaches the object with open fingers and then closes them to grasp the object. We employ a heuristic-based grasp for faster training. In real-world, the gripper fingers remain closed until the grasp mode is triggered. AnyGrasp predicts an appropriate grasp pose and the fingers close at the predicted position. To address the sim-to-real gap, we add randomization to the heuristic grasp pose during simulation. This allows the policy to generalize more effectively, resulting in more robust and reliable policies in the real-world.

\subsection{Domain Randomization Parameters}
\label{sec:domain_randomization}
To enhance the robustness of our policies for effective real-to-sim-to-real transfer, we apply domain randomization to various object properties and initial conditions. Table~\ref{tab:domain_randomization} details the key randomized parameters and their respective ranges. These variations ensure that our learned policies generalize effectively to real-world conditions, mitigating the discrepancies between simulation and real-world.

\begin{table}[h]
    \centering
    \resizebox{0.75\linewidth}{!}{ %
    \begin{tabular}{l c}
        \toprule
        \textbf{Parameter} & \textbf{Range} \\
        \midrule
        Object Scale & [0.8, 1.2] \\
        Mass & [0.3, 2.0] \\
        Friction & [0.3, 1.8] \\
        Restitution & [0.0, 1.0] \\
        Compliance & [0.0, 1.0] \\
        Center of Mass Perturbation & [-0.05, 0.05] \\
        Initial Position Perturbation & [-0.02, 0.02] \\
        Initial Orientation Perturbation & [-0.05, 0.05] \\
        Grasp Position Noise & [-0.01, 0.01] \\
        Grasp Orientation Noise & [-0.2, 0.2] \\
        \bottomrule
    \end{tabular}
    } %
    \caption{\small{Domain randomization ranges for key object properties and initial conditions in simulation.}}
    \label{tab:domain_randomization}
    \vspace{-2em}
\end{table}

\subsection{VLM Prompts}
\label{sec:prompt}
The VLM receives the image overlaid with keypoints ${1, \ldots, K}$, along with the task description as text. These are given to the VLM, along with the prompt. We do not provide any in-context examples with the prompt. Our prompt for single-step tasks is as follows:

\begin{lstlisting}
## Instructions
Your job is to help with moving rigid objects in real-world by writing code in python.
The task is given as an image of the environment, overlayed with keypoints marked with their indices, along with a text instruction.
These keypoints are in 3D space, and are projected onto the 2D image. They are attached with the objects, and move along with them.
So to determine where a specific point should go, you should specify where its corresponding keypoint should go.
The coordinate system is marked at the bottom right in the image, with a vertical arrow pointing forward in the positive x direction and the horizontal arrow pointing to the left in the positive y direction.
The code should predict the final keypoint locations relative to their matching keypoints. Use all matching keypoints to determine the final position of the moving object, not just a single reference point.
Note:
- You should determine if you need to grasp or push the object. You should output a boolean grasp_mode for that.
- Some objects should not be moved. Hence, the final location of key points on them should be the same as the initial locations.
- If you need to interact with an object, the final location of only the keypoints marked on it should change.
Hence, you should first try to understand which object should move.
- You should try to understand where the moving object should go relative to other stationary objects and use all the matching keypoints for alignment. Then you can give the final locations of keypoints of moving objects relative to keypoints on stationary objects.
- Positive x direction points towards up and positive y direction points towards left.
- The input to the function is a dictionary of keypoint coordinates. So keys will be strings like "1", "2", ... and their values will be numpy arrays ([x, y, z]) representing the 3D location of the keypoint corresponding to that index.
- To represent coordinates relative to other keypoints, you can make predictions like:
  keypoint_coordinates['1'] = keypoint_coordinates['2'] + np.array([delta_x, delta_y, delta_z]).
  For instance, if keypoint 1 needs to be placed to the left of keypoint 2, then delta_x = 0, delta_y = 0.1, and delta_z = 0.
  So can predict keypoint_coordinates['1'] = keypoint_coordinates['2'] + np.array([0, 0.1, 0]).
  The units here are in meters and left direction corresponds to + y-axis.
- Make use of semantics. Some objects should be placed in a certain way, like a left shoe should be placed on the left of the right shoe.

Structure your output in a single python code block as follows:

def get_interaction_data(keypoint_coordinates): 
    """ Put your explanation here. """ 
    object_to_interact = ? 
    keypoint_indices_to_interact = ? 
    grasp_mode = ? 
    ## final keypoint calculation for each keypoint in keypoint_indices_to_interact 
    keypoint_coordinates[`keypoint_indices_to_interact[0]`] = ? # Write calculation here. You may use multiple lines 
    # Repeat for other keypoints 
    return object_to_interact, keypoint_indices_to_interact, grasp_mode, keypoint_coordinates 

## Query 
Query Task: '[TASK]'
Query Image: [IMAGE_WITH_KEYPOINTS]
\end{lstlisting}

The prompt for multi-step tasks is as follows:

\begin{lstlisting}
## Instructions
Your job is to help with moving rigid objects in real-world by writing code in python.
The task is given as an image of the environment, overlayed with keypoints marked with their indices, along with a text instruction.
These keypoints are in 3D space, and are projected onto the 2D image. They are attached with the objects, and move along with them.
So to determine where a specific point should go, you should specify where its corresponding keypoint should go.
The coordinate system is marked at the bottom right in the image, with a vertical arrow pointing forward in the positive x direction and the horizontal arrow pointing to the left in the positive y direction.
The code should predict the final keypoint locations relative to their matching keypoints. Use all matching keypoints to determine the final position of the moving object, not just a single reference point.
Note:
- You should determine if you need to grasp or push the object. You should output a boolean grasp_mode for that. Generally, big objects can only be pushed as they are too big to be grasped.
- Some objects should not be moved. Hence, the final location of key points on them should be the same as the initial locations.
- If you need to interact with an object, the final location of only the keypoints marked on it should change.
  Hence, you should first try to understand which object should move.
- You should try to understand where the moving object should go relative to other stationary objects and use all the matching keypoints for alignment.
  Then you can give the final locations of keypoints of moving objects relative to keypoints on stationary objects.
- Positive x direction points towards up and positive y direction points towards left.
- The input to the function is a dictionary of keypoint coordinates. So keys will be strings like "1", "2", ... and their values will be numpy arrays ([x, y, z]) representing the 3D location of the keypoint corresponding to that index.
- To represent coordinates relative to other keypoints, you can make predictions like:
  keypoint_coordinates['1'] = keypoint_coordinates['2'] + np.array([delta_x, delta_y, delta_z]).
  For instance, if the keypoint 1 needs to be placed to the left of keypoint 2, then delta_x = 0, delta_y = 0.1, and delta_z = 0.
  So can predict keypoint_coordinates['1'] = keypoint_coordinates['2'] + np.array([0, 0.1, 0]).
  The units here are in meters and left direction corresponds to + y-axis.
- Make use of semantics. Some objects should be placed in a certain way, like a left shoe should be placed on the left of the right shoe.
- Some tasks involve multiple stages, so you will also predict the overall plan. Then you will write the description and code for the current stage. You will interact with only one object in a stage. Placing or pushing a single object will be considered a single stage. Grasping is not considered as a separate stage.
- You are free to make minor changes to the plan, or change the plan altogether if you think is necessary.
- We will keep adding the previous states of the environment as images and the corresponding code to the description. This will show how the task progressed. At the start, you will only see the task description.
- You should predict done=True when the task is complete, otherwise False. Only predict done=True when you see that the task is completed.

Structure your output in a single python code block as follows:

def get_interaction_data(keypoint_coordinates):
    """ 
        Previous Plan Description
        Current Plan Description
        Current stage description
    """
    done = ?
    if done:
        return
    object_to_interact = ?
    keypoint_indices_to_interact = ?
    grasp_mode = ?
    ## final keypoint calculation for each keypoint in keypoint_indices_to_interact
    keypoint_coordinates['keypoint_indices_to_interact[0]'] = ?  # Write calculation here. You may use multiple lines
    # Repeat for other keypoints
    return object_to_interact, keypoint_indices_to_interact, grasp_mode, keypoint_coordinates

## Query
Query Task: '[TASK]'
Query Image: [IMAGE_WITH_KEYPOINTS]
\end{lstlisting}

The prompt for baseline that uses pose input for single-step tasks is as follows:

\begin{lstlisting}
## Instructions
Your job is to help with moving rigid objects in a real-world environment by writing code in Python.
The task is given as an image of the environment, along with text instructions.
The coordinate system is marked at the bottom right in the image, with a vertical arrow pointing forward in the positive x direction and the horizontal arrow pointing to
the left in the positive y direction.
The objects are treated as rigid bodies and are labeled with numbers. You need to predict the final pose of the moving object relative to the pose of any object. It can be its own pose or pose of other objects in the image.
Note:
- You should determine if you need to grasp or push the object. You should output a boolean grasp_mode for that.
- Some objects should not be moved. Hence, the final pose of those objects will be the same as the initial pose.
- If you need to interact with an object, the final pose of that object should change.
- You should first try to understand which object should move relative to its pose, or poses of the other objects. Then you can give the final pose of the moving object relative to the other object poses.
- Positive x direction points upwards, and positive y direction points to the left.
- The input to the function is a dictionary of object poses. So the keys will be strings representing object labels like "1", "2", ..., and their values will be numpy arrays [x, y, z, r, p, y], where x, y, z represent the position in meters and r, p, y represent the orientation in radians.
- To represent poses relative to other objects, you can make predictions like: object_poses['1'] = object_poses['2'] + np.array([delta_x, delta_y, delta_z, delta_r, delta_p, delta_yaw]). For instance, if object 1 needs to be placed to the left of object 2, then: delta_x = 0, delta_y = 0.1, delta_z = 0, delta_r = 0, delta_p = 0, delta_yaw = 0. So, you can predict: object_poses['1'] = object_poses['2'] + np.array([0, 0.1, 0, 0, 0, 0]). The units for positions are in meters, and for orientations, they are in radians. The left direction corresponds to the positive y-axis.
- Make use of semantics. Some objects should be placed in a certain way, like a left shoe should be placed on the left of the right shoe.
Structure your output in a single Python code block as follows:

def get_final_poses(object_poses):  
    """ Put your explanation here. """  
    object_to_interact = ?  
    grasp_mode = ?  
    ## final pose calculation for each object  
    object_poses[`object_to_interact`] = ?  # Write calculation here. You may use multiple lines  
    return object_to_interact, grasp_mode, object_poses  

## Query 
Query Task: '[TASK]'
Query Image: [IMAGE]
\end{lstlisting}

\subsection{Case study of a complex task}
\begin{figure}[h]
    \setlength{\abovecaptionskip}{0pt}
    \includegraphics[width=\linewidth]{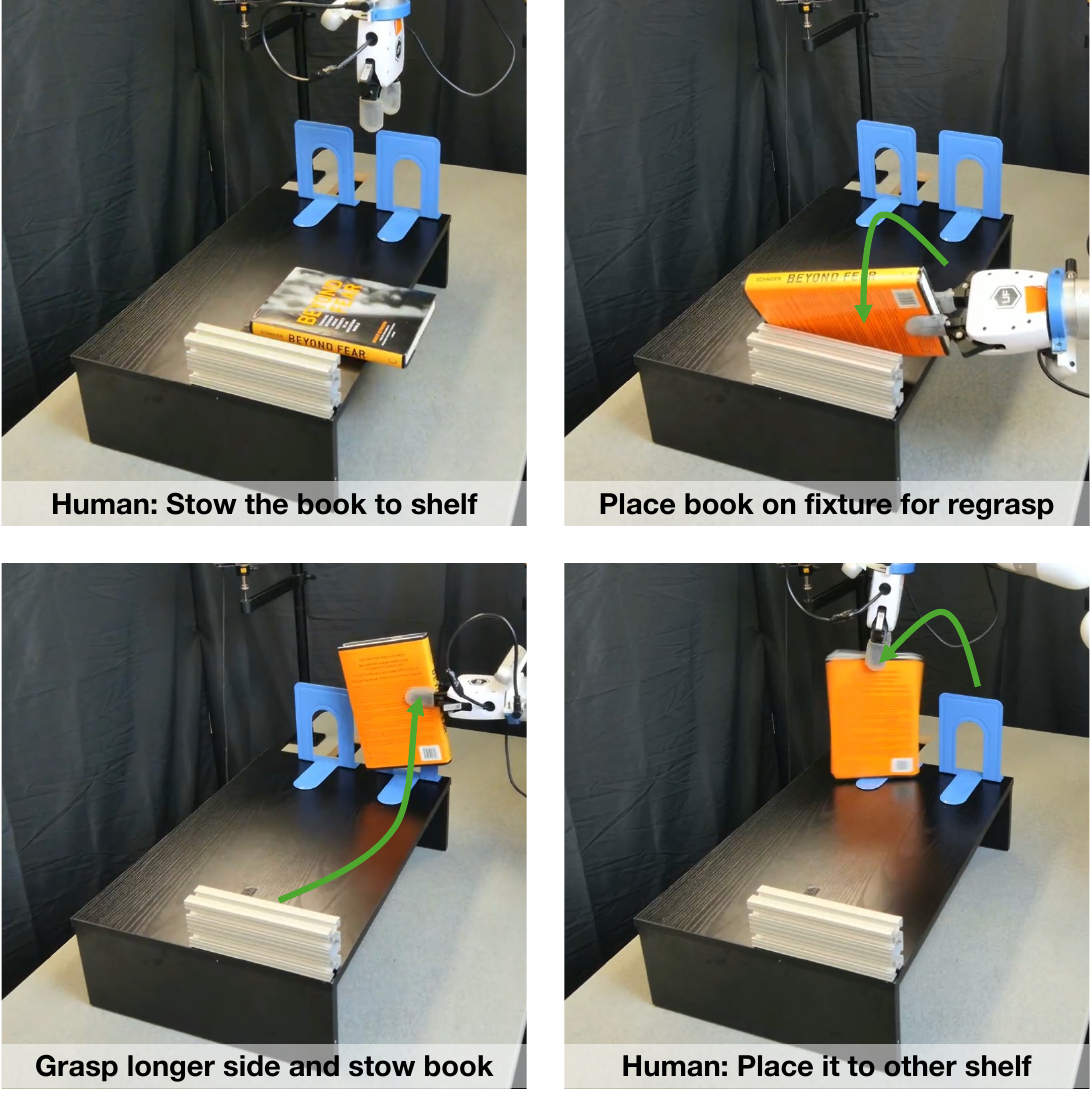}
    \caption{\small{\textbf{Case study of a complex task with in-context examples.} The robot uses the environment to regrasp and stow the book. Then, the human updates the instructions to place it on the other shelf.}}
    \label{fig:case_study}
\end{figure}

We present results on a complex 3D understanding task. The task involves stowing a book on a shelf, where the book is initially positioned with only its shorter edge graspable. The instruction is to place the book on the shelf. However, the robot cannot place the book directly with the shorter edge grasped, as this would result in a collision between the book and the table due to the position of its arm. To complete this task, the robot must perform multiple steps: first, it needs to regrasp the book along its longer edge using some part of the environment, and only then can it stow the book on the shelf. After the robot places the book on the initial shelf, a human intervenes by adding an instruction to move the book to a different shelf.

Given the complexity of this long-horizon task, we employ in-context examples to guide the VLM. With this change,  our system is able to successfully perform the task. \figref{fig:case_study} illustrates the progression of the task.

\end{document}